%% file: bare_jrnl_new_sample4.tex
\documentclass[lettersize,journal]{IEEEtran}
\usepackage{amsmath,amsfonts}
\usepackage{algorithmic}
\usepackage{algorithm}
\usepackage{array}
\usepackage[caption=false,font=normalsize,labelfont=sf,textfont=sf]{subfig}
\usepackage{textcomp}
\usepackage{stfloats}
\usepackage{url}
\usepackage{verbatim}
\usepackage{graphicx}
\usepackage{cite}
\usepackage{multirow}
\usepackage{rotating}
\usepackage{lettrine}
\usepackage{hyperref}

\begin{document}

\title{CWD30: A Comprehensive and Holistic Dataset for Crop Weed Recognition in Precision Agriculture}

\author{\IEEEauthorblockN{Talha Ilyas\IEEEauthorrefmark{1}\IEEEauthorrefmark{2},
		Dewa Made Sri Arsa\IEEEauthorrefmark{1}\IEEEauthorrefmark{3},
		Khubaib Ahmad\IEEEauthorrefmark{1}\IEEEauthorrefmark{2}, 
		Yong Chae Jeong\IEEEauthorrefmark{1},
		Okjae Won\IEEEauthorrefmark{4},
		Jong Hoon Lee\IEEEauthorrefmark{2}, and
		Hyongsuk Kim\IEEEauthorrefmark{2}
	} \\
	\IEEEauthorblockA{\IEEEauthorrefmark{1}Division of Electronics and Information Engineering,
		Jeonbuk National University, Jeonju 54896, Republic of Korea}
	
	\IEEEauthorblockA{\IEEEauthorrefmark{2}Core Research Institute of Intelligent Robots, Jeonbuk National University, Jeonju 54896, Republic of Korea}
	
	\IEEEauthorblockA{\IEEEauthorrefmark{3}Department of Information Engineering, Universitas Udayana, Bali, Indonesia}
	
	\IEEEauthorblockA{\IEEEauthorrefmark{4}Production Technology Research Division, Rural Development Administration, National Institute of Crop Science, Miryang, Republic of Korea}

}



\maketitle

\begin{abstract}
The growing demand for precision agriculture necessitates efficient and accurate crop-weed recognition and classification systems. Current datasets often lack the sample size, diversity, and hierarchical structure needed to develop robust deep learning models for discriminating crops and weeds in agricultural fields. Moreover, the similar external structure and phenomics of crops and weeds complicate recognition tasks. To address these issues, we present the CWD30 dataset, a large-scale, diverse, holistic, and hierarchical dataset tailored for crop-weed recognition tasks in precision agriculture. CWD30 comprises over 219,770 high-resolution images of 20 weed species and 10 crop species, encompassing various growth stages, multiple viewing angles, and environmental conditions. The images were collected from diverse agricultural fields across different geographic locations and seasons, ensuring a representative dataset. The dataset's hierarchical taxonomy enables fine-grained classification and facilitates the development of more accurate, robust, and generalizable deep learning models. We conduct extensive baseline experiments to validate the efficacy of the CWD30 dataset. Our experiments reveal that the dataset poses significant challenges due to intra-class variations, inter-class similarities, and data imbalance. Additionally, we demonstrate that minor training modifications like using CWD30 pretrained backbones can significantly enhance model performance and reduce convergence time, saving training resources on several downstream tasks. These challenges provide valuable insights and opportunities for future research in crop-weed detection, fine-grained classification, and imbalanced learning. We believe that the CWD30 dataset will serve as a benchmark for evaluating crop-weed recognition algorithms, promoting advancements in precision agriculture, and fostering collaboration among researchers in the field.
The data is available at:  \url{https://github.com/Mr-TalhaIlyas/CWD30}
\end{abstract}


\input{introduction}
\input{relatedworks}
\input{data}
\input{experiments}
\input{results}

\section{Conclusion}
In conclusion, this paper presents the CWD30 dataset, a comprehensive, holistic, large-scale, and diverse crop-weed recognition dataset tailored for precision agriculture. With over 219,770 high-resolution images of 20 weed species and 10 crop species, the dataset spans various growth stages, multiple viewing angles, and diverse environmental conditions. The hierarchical taxonomy of CWD30 facilitates the development of accurate, robust, and generalizable deep learning models for crop-weed recognition. Our extensive baseline experiments demonstrate the challenges and opportunities presented by the CWD30 dataset. These experiments emphasize the importance of utilizing CWD30 pretrained backbones, which result in enhanced performance, reduced convergence time, and consequently, saved time and training resources for various fine-tuning and downstream precision agriculture tasks. The CWD30 dataset not only advances research in the field of precision agriculture but also promotes collaboration among researchers by serving as a benchmark for evaluating crop-weed recognition algorithms.


\begin{table*}[]
	\centering
	\caption{Detailed Taxonomy of Plant Species Included in the CWD30 Dataset. The kingdom and phylum of all plants listed are plantae and magnoliophyta respectively.}
	\label{table-appendix}
	\begin{tabular}{llllllll}
		\hline
		\textbf{Common   Name} & \textbf{Scientific Name} & \textbf{Order} & \textbf{Family} & \textbf{Genus} & \textbf{Species} & \textbf{Class} & \textbf{Sub-Class} \\ \hline
		Asian flatsedge & Cyperus microiria & Poales & Cyperaceae & Cyperus & microiria & Weed & broad-leaves \\
		Asiatic dayflower & Commelina communis & Commelinales & Commelinaceae & Commelina & communis & Weed & broad-leaves \\
		Bean & Phaseolus vulgaris & Fabales & Fabaceae & Phaseolus & vulgaris & Crop & legumes \\
		Bloodscale sedge & Carex haematostoma & Poales & Cyperaceae & Carex & haematostoma & Weed & sedge \\
		Cockspur grass & Echinochloa crus-galli & Poales & Poaceae & Echinochloa & crus-galli & Weed & grass \\
		Copperleaf & Acalypha spp. & Malpighiales & Euphorbiaceae & Acalypha & spp. & Weed & broad-leaves \\
		Corn & Zea mays & Poales & Poaceae & Zea & mays & Crop & grains \\
		Early barnyard grass & Echinochloa oryzoides & Poales & Poaceae & Echinochloa & oryzoides & Weed & grass \\
		Fall panicum & Panicum dichotomiflorum & Poales & Poaceae & Panicum & dichotomiflorum & Weed & grass \\
		Finger grass & Digitaria sanguinalis & Poales & Poaceae & Digitaria & sanguinalis & Weed & grass \\
		Foxtail millet & Setaria italica & Poales & Poaceae & Setaria & italica & Crop & grains \\
		Goosefoot & Chenopodium album & Caryophyllales & Amaranthaceae & Chenopodium & album & Weed & broad-leaves \\
		Great millet & Sorghum bicolor & Poales & Poaceae & Sorghum & bicolor & Crop & grains \\
		Green foxtail & Setaria viridis & Poales & Poaceae & Setaria & viridis & Weed & grass \\
		Green gram & Vigna radiata & Fabales & Fabaceae & Vigna & radiata & Crop & legumes \\
		Henbit & Lamium amplexicaule & Lamiales & Lamiaceae & Lamium & amplexicaule & Weed & broad-leaves \\
		Indian goosegrass & Eleusine indica & Poales & Poaceae & Eleusine & indica & Weed & grass \\
		Korean dock & Rumex crispus & Caryophyllales & Polygonaceae & Rumex & crispus & Weed & broad-leaves \\
		Livid pigweed & Amaranthus lividus & Caryophyllales & Amaranthaceae & Amaranthus & lividus & Weed & broad-leaves \\
		Nipponicus sedge & Carex nipponica & Poales & Cyperaceae & Carex & nipponica & Weed & sedge \\
		Peanut & Arachis hypogaea & Fabales & Fabaceae & Arachis & hypogaea & Crop & broad-leaves \\
		Perilla & Perilla frutescens & Lamiales & Lamiaceae & Perilla & frutescens & Crop & oil seeds \\
		Poa annua & Poa annua & Poales & Poaceae & Poa & annua & Weed & grasses \\
		Proso millet & Panicum miliaceum & Poales & Poaceae & Panicum & miliaceum & Crop & grains \\
		Purslane & Portulaca oleracea & Caryophyllales & Portulacaceae & Portulaca & oleracea & Weed & broad-leaves \\
		Red bean & Phaseolus angularis & Fabales & Fabaceae & Phaseolus & angularis & Crop & broad-leaves \\
		Redroot pigweed & Amaranthus retroflexus & Caryophyllales & Amaranthaceae & Amaranthus & retroflexus & Weed & broad-leaves \\
		Sesame & Sesamum indicum & Lamiales & Pedaliaceae & Sesamum & indicum & Crop & oil seeds \\
		Smooth pigweed & Amaranthus hybridus & Caryophyllales & Amaranthaceae & Amaranthus & hybridus & Weed & broad-leaves \\
		White goosefoot & Chenopodium album & Caryophyllales & Amaranthaceae & Chenopodium & album & Weed & broad-leaves \\ \hline
	\end{tabular}
\end{table*}

\section*{Acknowledgments}
This work was supported in part by the Agricultural Science and Technology Development Cooperation Research Program (PJ015720) and Basic Science Research Program through the National Research Foundation of Korea (NRF) funded by the Ministry of Education (NRF-2019R1A6A1A09031717 and NRF-2019R1A2C1011297).

\appendix[Taxonomy of plant species]
See Table \ref{table-appendix}.

\bibliographystyle{IEEEtran}
\bibliography{citation}

\newpage

\vspace{11pt}

\vfill

\end{document}

%% file: introduction.tex
\section {Introduction}
\lettrine{P}{recision} agriculture is essential to address the increasing global population and the corresponding demand for a 70\% increase in agricultural production by 2050 \cite{radoglou2020compilation}. The challenge lies in managing limited cultivation land, water scarcity, and the effects of climate change on productivity. One critical aspect of precision agriculture is the effective control of weeds that negatively impact crop growth and yields by competing for resources and interfering with crop growth through the release of chemicals \cite{iqbal2019investigation,patel2016weed,moazzam2019review}. 

Recent advances in deep learning have revolutionized the field of computer vision, with Convolutional Neural Networks (CNNs) and transformers becoming the backbone of numerous state-of-the-art models \cite{ilyas2022diana,elsherbiny2021integration,sa2017weednet}. However, their performance relies heavily on the quality and diversity of training data \cite{hasan2021survey,bai2021transformers}, emphasizing the importance of comprehensive agricultural datasets for model development \cite{joshi2022standardizing}. But the agricultural domain often suffers from the deficiency of task-specific data \cite{shorten2019survey1,joshi2022standardizing}. Which can result in insufficient data variety, overfitting, inadequate representation of real-world challenges, and reduced model robustness. These limitations hinder the model's ability to generalize and accurately recognize crops and weeds in diverse real-world situations. To overcome these issues, researchers employ techniques like data augmentation \cite{su2021data,shorten2019survey2}, transfer learning \cite{espejo2020towards} , or synthetic data generation \cite{cap2020leafgan}, although these approaches may not always achieve the same performance level as models trained on larger, more diverse datasets \cite{moon2021knowledge}. Transfer learning (fine-tuning) \cite{pan2010survey} is a common approach for training deep learning models in agriculture, as it involves using pretrained weights from other tasks (e.g., ImageNet) to address data deficiency \cite{antonijevic2023transfer}. Pretrained weights from ImageNet \cite{deng2009imagenet} and COCO \cite{lin2014microsoft} are commonly used but are less suitable for domain-specific agricultural tasks due to their generic content \cite{agrinet,joshi2022standardizing}. Thus absence of a centralized benchmark repository for agriculture-specific datasets hinders the development of computer-aided precision agriculture (CAPA) systems. 

\begin{figure*}[!t]
	\centering
	\includegraphics[width=\textwidth]{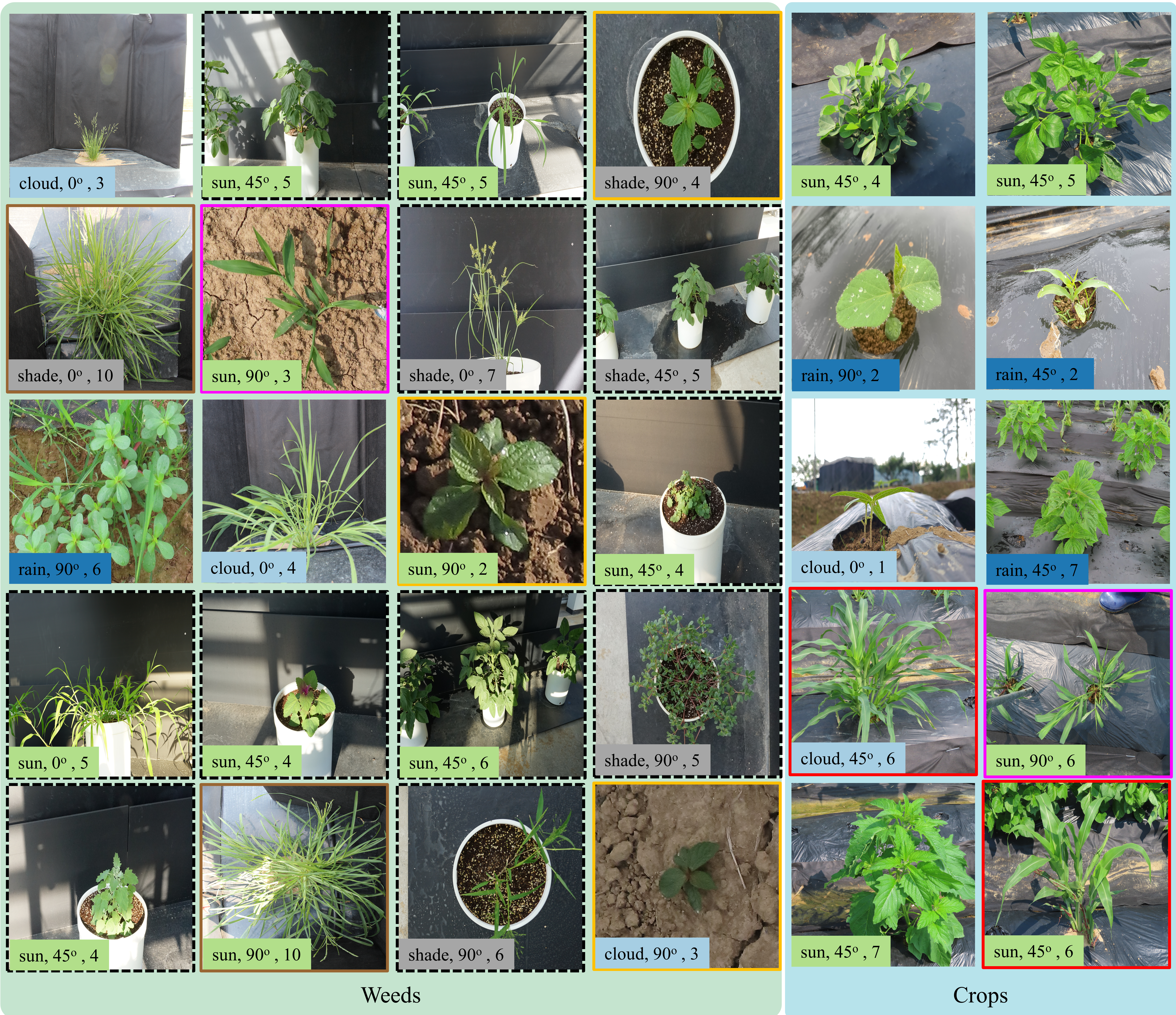}
	\caption{Crop and Weed image samples from CWD30 dataset, captured at different life cycle stages, under varying environment and from different viewing angles.Key elements in the images are highlighted: pink-bordered images represent similarities at a macro class level (crop vs weed); orange boxes indicate the variability within a single weed species due to environmental factors such as indoor vs outdoor settings and soil type; images encased in red and brown borders demonstrate visually similar crop and weed classes; images marked with black dashed lines represent weeds cultivated in a laboratory setting; small inset boxes on each image provide information about the weather conditions and camera angle and plant age at time of capture.}
	\label{fig1}
\end{figure*}

\begin{figure}
	\centering
	\includegraphics[width=3in]{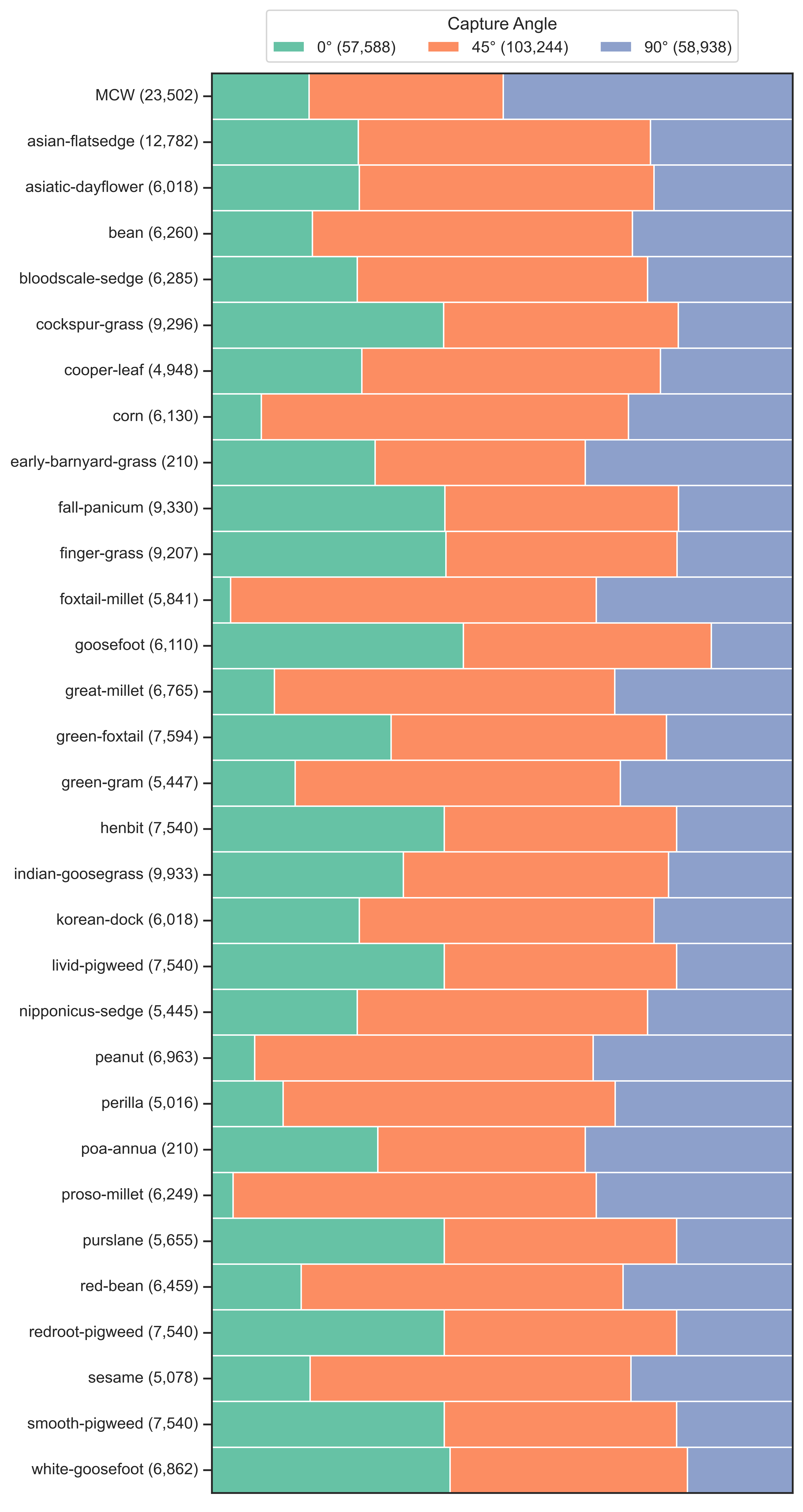}
	\caption{A comparative plot of class distributions per viewing angle. Numbers in parenthesis represent the total number of images of that plant category.}
	\label{fig2}
\end{figure}

In this study, we introduce and evaluate the crop weed recognition dataset (CWD30) dataset, a large-scale and diverse collection of various crops and weed images that captures the complexities and challenges of real-world precision agriculture scenarios. The CWD30 dataset comprises a collection of 219,770 images that encompass 10 crop classes and 20 weed classes. These images capture various growth stages, multiple viewing angles, and diverse environmental conditions. Figure \ref{fig1} shows some image samples, while Figure \ref{fig2} displays the number of images per category. The CWD30 dataset addresses the significant intra-class difference and large inter-species similarity of multiple crop and weed plants. We train various deep learning models, including CNNs and transformer-based architectures, on the CWD30 dataset to assess their performance and investigate the impact of pretraining. Furthermore, we analyze the structure of the feature embeddings obtained by these models and compare their performance on downstream tasks, such as pixel-level crop weed recognition

In summary, building upon the aforementioned challenges and limitations we make the following main contributions:
\begin{itemize}
	\item We present the crop-weed dataset (CWD30), which, to the best of our knowledge, is the first truly holistic, large-scale crop weed recognition dataset available to date.
	\item	Proposed dataset encompasses a wide range of plant growth stages, i.e., from seedlings to fully mature plants. This extensive coverage of growth stages ensures that the CWD30 dataset captures the various morphological changes and developmental stages plants undergo throughout their life cycle. By incorporating these diverse growth stages, the dataset provides a more comprehensive representation of real-world agricultural scenarios. Consequently, deep learning models trained on this dataset can better adapt to the inherent variability in plant appearances and growth stages, Figure \ref{fig7} shows a few samples of plants at various growth stages.
	\item The CWD30 dataset offers a unique advantage by including multi-view images, captured at various angles. This comprehensive representation of plants account for various viewpoints and lighting conditions, which enhances the dataset's ability to model real-world situations. The multi-view images enable the development of more robust and generalizable deep learning models, as they allow the models to learn from a broader range of visual features and better understand the complexities and variations commonly found in real-field settings (see section III for details).
	\item	Compared to existing agricultural datasets that focus on specific plant parts like branches or leaves, the proposed CWD30 dataset offers high-resolution images of entire plants in various growth stages and viewpoints. This comprehensive nature of the CWD30 dataset allows for the generation of simpler, plant-part-specific datasets by cropping its high-resolution images. As a result, the CWD30 dataset can be considered a more versatile and complete resource compared to existing datasets. This dataset contributes to overcoming the limitations of previous datasets and advances the field of precision agriculture.
	\item	Additionally, we demonstrate that models pretrained on the CWD30 dataset consistently outperform their ImageNet-1K pretrained counterparts, yielding more meaningful and robust feature representations. This improvement, in turn, enhances the performance of state-of-the-art models on popular downstream agricultural tasks (see section V for details).
\end{itemize}
These contributions can further advance the research and development of reliable CAPA systems.

The rest of this article unfolds as follows: Section II provides a review of related literature and relevant datasets. Section III explains the development of the CWD30 dataset, its unique characteristics, and draws comparisons with other agricultural datasets. The experimental setup is outlined in Section IV. Following this, Section V delves into the analysis of experimental results and the inherent advantages offered by the CWD30 dataset. Finally, we wrap up the article in the conclusion.

\begin{figure}
	\centering
	\includegraphics[width=3in]{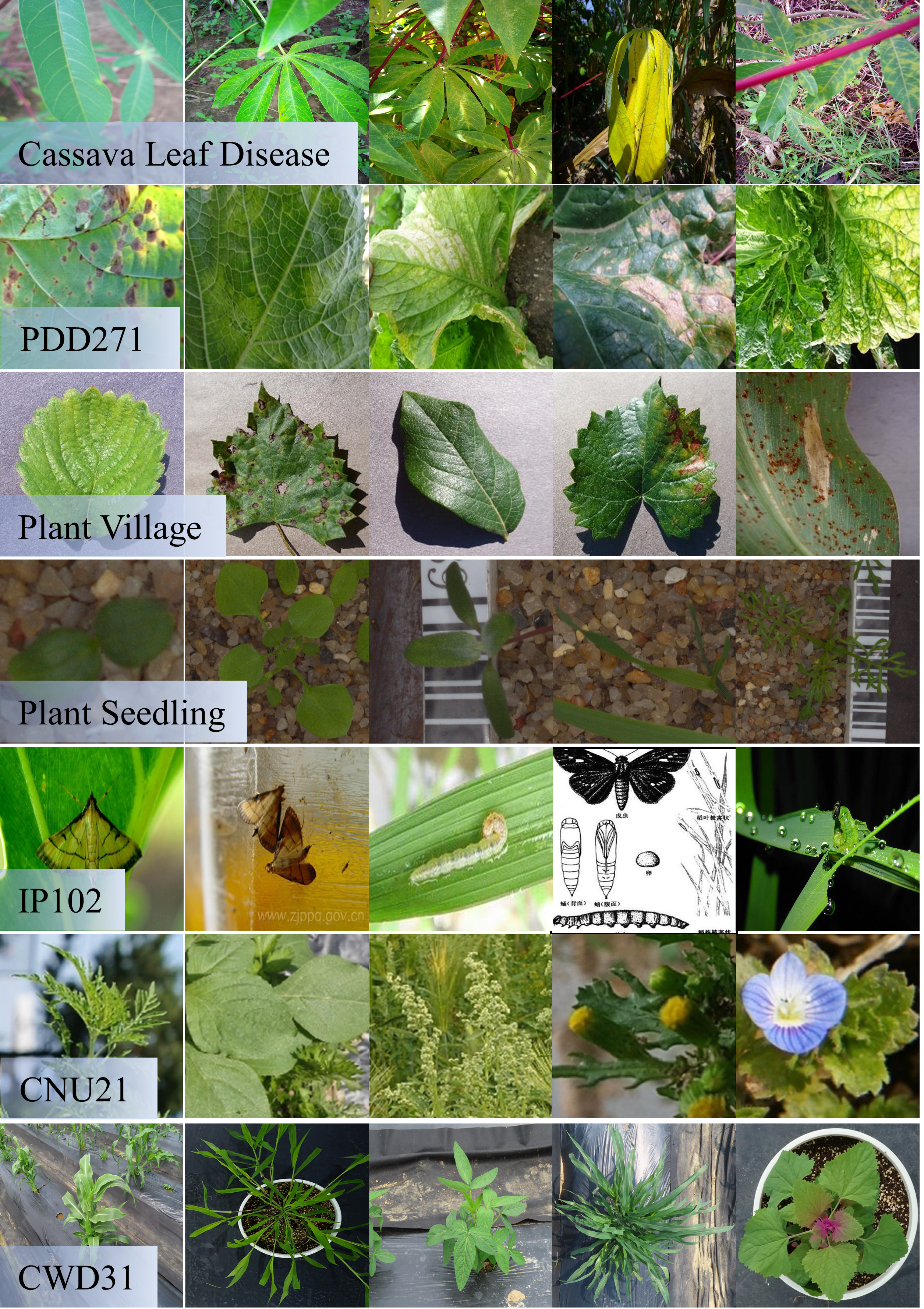}
	\caption{Visual comparison of CWD30 dataset with other related datasets.}
	\label{fig3}
\end{figure}

%% file: relatedworks.tex
\begin{figure*}[!t]
	\centering
	\includegraphics[width=\textwidth]{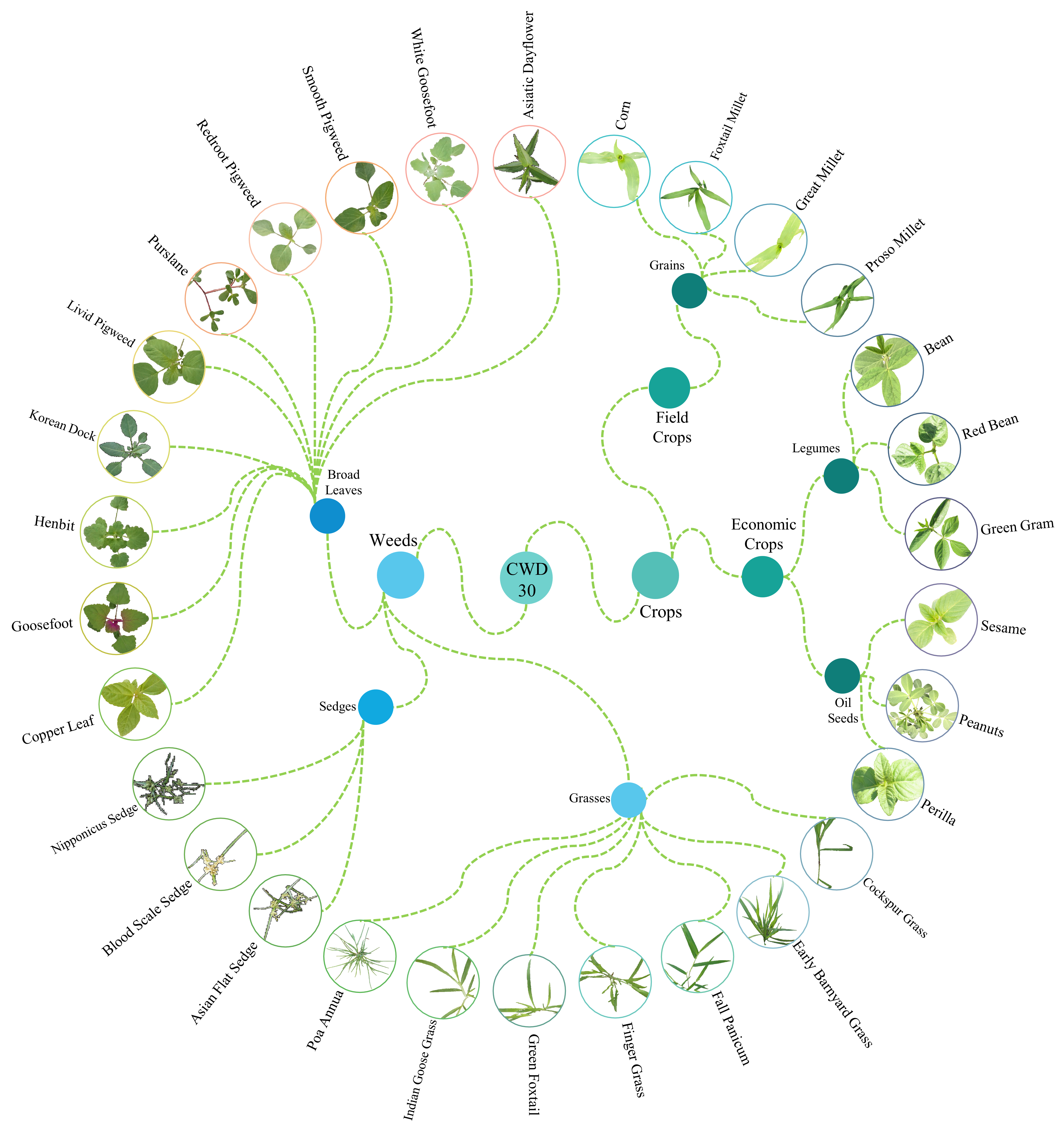}
	\caption{Taxonomy of CWD30 dataset. Showcasing the hierarchical organization of crop and weed species included in the dataset.}
	\label{fig4}
\end{figure*}

\section {Related Works}
\subsection{Crop Weed Recognition}
Crop-weed recognition is vital in CAPA systems for efficient and sustainable farming practices. Reliable recognition and differentiation allow for effective weed management and optimal crop growth, reducing chemical usage and minimizing environmental impact \cite{kamilaris2018deep,hasan2021survey}. It also helps farmers oversee their crops' health, enabling prompt response and lowering the possibility of crop loss from weed infestations \cite{fuentes2019deep,ilyas2022diana}. However, these systems face limitations due to the reliance on small datasets \cite{rahnemoonfar2017deep}, resulting in reduced model robustness, overfitting, and inadequate representation of real-world challenges.

\begin{sidewaystable}
	\caption{Comparative Analysis of Various Agricultural Datasets: Key Attributes and Characteristics.The symbol `\textasciitilde' indicates an approximate value. HH, DM, and VM correspond to handheld, device mounted, and vehicle mounted cameras, respectively.}
	\label{tab1}
	\resizebox{\textwidth}{!}{
		\begin{tabular}{lllllllllllll}
			\hline
			\multicolumn{1}{c}{\textbf{Dataset}} & \multicolumn{1}{c}{\textbf{\# Images}} & \multicolumn{1}{c}{\textbf{\# Cat.}} & \multicolumn{1}{c}{\textbf{Coverage}} & \multicolumn{1}{c}{\textbf{Environment}} & \multicolumn{1}{c}{\textbf{Background}} & \multicolumn{1}{c}{\textbf{\begin{tabular}[c]{@{}c@{}}Avg. Image\\ Resolution\end{tabular}}} & \multicolumn{1}{c}{\textbf{\begin{tabular}[c]{@{}c@{}}Multi\\ View\end{tabular}}} & \multicolumn{1}{c}{\textbf{\begin{tabular}[c]{@{}c@{}}Growth\\ Stages\end{tabular}}} & \multicolumn{1}{c}{\textbf{Availability}} & \multicolumn{1}{c}{\textbf{Image Content}} & \multicolumn{1}{c}{\textbf{Location}} & \multicolumn{1}{c}{\textbf{\begin{tabular}[c]{@{}c@{}}Acquisition\\ Platform\end{tabular}}} \\ \hline
			Deep Weeds \cite{olsen2019deepweeds} & 17,509 & 9 & weeds & outdoor & complex & 256x256 & \multirow{3}{*}{No} & \multirow{3}{*}{No} & public & Full plant & Roadside & \multirow{3}{*}{\begin{tabular}[c]{@{}l@{}}Tripod / \\ Overhead\\ Camera\end{tabular}} \\
			Plant Seedling \cite{giselsson2017public} & 5,539 & 12 & weeds & indoor & simple & $\sim$355x355 &  &  & public & Full plant & Trays &  \\
			Fruit Leaf \cite{chouhan2019data} & 4,503 & 12 & fruits & indoor & simple & 6000x4000 &  &  & public & single leaf & tray &  \\ \hline
			PDDB \cite{barbedo2019plant} & 46,409 & 56 & Fruits, crops & indoor & simple & 2048x1368 & \multirow{3}{*}{No} & \multirow{3}{*}{No} & public & \multirow{3}{*}{Single leaf} & \multirow{3}{*}{Lab} & \multirow{10}{*}{\begin{tabular}[c]{@{}l@{}}Handheld\\ RGB\\ Camera\end{tabular}} \\
			Corn2022 \cite{qian2022deep} & 7,701 & 4 & corn & outdoor & Simple & 224x224 &  &  & public &  &  &  \\
			LWDCD2020 \cite{goyal2021leaf} & 12,160 & 10 & wheat & outdoor & simple & 224x224 &  &  & private &  &  &  \\ \cline{1-12}
			Plant Village \cite{hughes2015open} & 54,309 & 38 & fruits, crops & indoor & simple & 256x256 & \multirow{2}{*}{No} & \multirow{2}{*}{No} & \multirow{2}{*}{public} & \multirow{2}{*}{Single leaf} & \multirow{2}{*}{Lab} &  \\
			Plant Doc \cite{singh2020plantdoc} & 2,598 & 17 & fruits, crops & outdoor & complex & $\sim$1070x907 &  &  &  &  &  &  \\ \cline{1-12}
			RiceLeaf \cite{sethy2020deep} & 5,932 & 4 & rice & outdoor & simple & 300x300 & \multirow{5}{*}{No} & \multirow{5}{*}{No} & private & Single leaf & \multirow{5}{*}{Farmland} &  \\
			CLD \cite{ayu2021deep} & 15,000 & 6 & cassava & outdoor & complex & 800x600 &  &  & public & Single branch &  &  \\
			AppleLeaf \cite{thapa2020plant} & 23,249 & 6 & frutis & outdoor & simple & 4000x2672 &  &  & public & Single leaf &  &  \\
			CNU \cite{trong2020late} & 208,477 & 21 & weeds & outdoor & complex & - &  &  & private & Single branch &  &  \\
			PDD271 \cite{liu2021plant} & 220,592 & 271 & \begin{tabular}[c]{@{}l@{}}fruits, crops,\\ vegetables\end{tabular} & outdoor & Simple & 256x256 &  &  & private & Single leaf &  &  \\ \hline
			IP102 \cite{wu2019ip102} & 75,222 & 102 & crop pests & \begin{tabular}[c]{@{}l@{}}Simple /\\ complex\end{tabular} & Simple & $\sim$525x413 & No & No & private & \begin{tabular}[c]{@{}l@{}}Single pest\\ on leaf\end{tabular} & \begin{tabular}[c]{@{}l@{}}Farmland, sketch,\\ drawings\end{tabular} & \begin{tabular}[c]{@{}l@{}}Search\\ Engines\end{tabular} \\ \hline
			\textbf{CWD30} & 219,778 & 30 & crops, weeds & \begin{tabular}[c]{@{}l@{}}Simple /\\ complex\end{tabular} & \begin{tabular}[c]{@{}l@{}}Simple /\\ complex\end{tabular} & $\sim$4032x3024 & Yes & Yes & public & Full plant & Farmland, Pots & \begin{tabular}[c]{@{}l@{}}HH /DM / \\ VM / Overhead\\ camera\end{tabular} \\ \hline
		\end{tabular}%
	}
\end{sidewaystable}

Several studies have shown the potential of deep learning techniques in addressing key components and challenges in developing CAPA systems, such as unmanned weed detection \cite{arsa2023eco}, fertilization \cite{escalante2019barley}, irrigation, and phenotyping \cite{yi2020deep}. Kamilaris et al. \cite{kamilaris2018deep} conducted experiments that showed deep learning outperforming traditional methods. Westwood et al. \cite{westwood2018weed} discussed the potential of deep learning-based plant classification for unmanned weed management. Wang et al.  highlighted the main challenges in differentiating weed and crop species in CAPA systems. Moreover, Wang et al. \cite{wang2019review} and Khan et al. \cite{khan2022systematic} emphasized the importance of combining spectral and spatial characteristics for remote sensing and ground-based weed identification approaches. Hasan et al. \cite{hasan2021survey} conducted a comprehensive survey of deep learning techniques for weed detection and presented a taxonomy of deep learning techniques.

However, recent studies by Moazzam et al. \cite{moazzam2019review} and Coleman et al. \cite{coleman2023more} identified research gaps, such as a lack of substantial crop-weed datasets and generalized models and concluded that methods like data augmentation and transfer learning might not always produce results on par with models trained on more substantial, diverse datasets. To address these limitations and challenges, further research is needed to improve the accuracy and robustness of CAPA systems. Considering the identified research gaps and challenges, this work presents the CWD30 dataset, specifically designed to address the limitations of existing agricultural datasets. Our aim is to facilitate the development of accurate and reliable CAPA systems, ultimately enhancing the effectiveness and sustainability of precision agriculture practices.

\begin{table*}[!t]
	\centering
	\caption{List of weed species included in the CWD30 dataset, their geographical distribution, and the crop species they commonly affect, emphasizing the importance of weed recognition and management in global agriculture \cite{USDAPlants,WSSA}.}
	\label{tab2}
	\resizebox{6in}{!}{%
		\begin{tabular}{llllllllll}
			\cline{1-3}
			\multicolumn{1}{c}{\textbf{Weed Name}} & \multicolumn{1}{c}{\textbf{Countries Found In}} & \multicolumn{1}{c}{\textbf{Crops Affected}} &  \\ \cline{1-3}
			Cockspur   Grass & United   States, Canada, Europe, Asia, Australia, Africa & Corn,   millets &  \\
			Early   Barnyard Grass & Europe,   Asia, Africa & Corn,   millets &  \\
			Fall   panicum & North   America, Europe, Asia & Corn,   millets &  \\
			Fingergrass & Worldwide & Corn,   millets &  \\
			Green   foxtail & North   America, Europe, Asia & Corn,   millets &  \\
			Indian   goosegrass & Asia,   Africa, South America & Corn,   millets &  \\
			Poa   annua & Worldwide & Corn,   millets &  \\
			Copper   leaf & Worldwide & Corn,   millets, beans, peanuts &  \\
			Goosefoot & Worldwide & Corn,   millets, beans &  \\
			Henbit & North   America, Europe, Asia & Corn,   millets, beans &  \\
			Livid   pigweed & North   America, Europe, Asia & Corn,   millets, beans &  \\
			Purslane & Worldwide & Corn,   millets, beans &  \\
			Redroot   pigweed & North   America, Europe, Asia & Corn,   millets, beans &  \\
			Smooth   pigweed & North   America, Europe, Asia & Corn,   millets, beans &  \\
			White   goosefoot & North   America, Europe, Asia & Corn,   millets, beans &  \\
			Asian   flats edge & Asia,   North America, South America & Millets,   beans &  \\
			Bloodscale   sedge & North   America, Europe, Asia & Millets &  \\
			Nipponicus   sedge & Asia,   Europe, North America & Millets &  \\
			Korean   dock & Asia,   North America, Europe & Millets,   beans, sesame &  \\
			Asiatic   dayflower & Asia,   North America, Europe, Australia & Millets,   beans, sesame &  \\ \cline{1-3}
		\end{tabular}%
	}
\end{table*}

\subsection {Related Datasets}
Here we provide an overview of several related agricultural datasets that have been previously proposed for crop-weed recognition and other agricultural tasks \cite{olsen2019deepweeds,giselsson2017public,chouhan2019data,barbedo2019plant, qian2022deep,goyal2021leaf,hughes2015open,singh2020plantdoc,sethy2020deep,ayu2021deep,thapa2020plant,trong2020late, liu2021plant,wu2019ip102}. These datasets, while valuable, have certain limitations that the CWD30 dataset aims to address.

\textbf{Plant Seedling:}The Plant Seedlings Dataset \cite{giselsson2017public} features approximately 960 unique plants from 11 species at various growth stages. It consists of annotated RGB images with a resolution of around 10 pixels per mm. Three public versions of the dataset are available: original, cropped, and segmented. For comparison in this study, we use the cropped plants v2 version, which contains 5,539 images of 12 different species. The dataset is imbalanced, with some classes having up to 654 samples (chickweed) and others as few as 221 (wheat).

The dataset was collected over 20 days (roughly 3 weeks) at 2-to-3-day intervals in an indoor setting. Plants were grown in a styrofoam box, and images were captured using a fixed overhead camera setup. This database was recorded at the Aarhus University Flakkebjerg Research station as part of a collaboration between the University of Southern Denmark and Aarhus University.

\textbf{CNU:} This weeds dataset from Chonnam National University (CNU) in the Republic of Korea \cite{trong2020late} consists of 208,477 images featuring 21 species. Captured on farms and fields using high-definition cameras, the images encompass various parts of weeds, including flowers, leaves and fruits. A visual comparison between the CNU dataset and the CWD30 dataset is illustrated in a Figure \ref{fig3}. However, unlike the CWD30 dataset, the CNU dataset does not encompass the growth stages and multiple viewing angles. The CNU dataset is imbalanced, with over 24,300 images of shaggy soldier and only about 800 images of spanish needles.

\begin{table*}
	\centering
	\caption{Global production share ,in million metric tons (M), of the 10 crop species included in the CWD30 dataset for the year 2020 to 2021, across various countries, emphasizing their significance and contribution to worldwide agricultural production \cite{RDAKorea,USDAProduction,leff2004geographic}.}
	\label{tab3}
	\renewcommand{\arraystretch}{1.5}
	\resizebox{6in}{!}{%
		\begin{tabular}{lccccccccc}
			\hline
			\textbf{Country} & \textbf{Corn} & \textbf{Foxtail Millet} & \textbf{Great Millet} & \textbf{Proso Millet} & \textbf{Bean} & \textbf{Green Gram} & \textbf{Peanut} & \textbf{Red Bean} & \textbf{Sesame} \\ \hline
			\textbf{United States} & 358.4M & - & 9.7M & - & - & - & 2.79M & - & - \\
			\textbf{China} & 260.8M & 6.5M & - & 1.8M & - & 0.6M & 17.9M & 2.2M & - \\
			\textbf{Brazil} & 81M & - & - & - & 4.2M & - & - & - & - \\
			\textbf{India} & 31.65 & 1M & 6M & 2.2M & 6.5M & 2M & 6.7M & - & 0.8M \\
			\textbf{Nigeria} & 12.4 & - & 7.1M & - & - & - & 4.23M & - & - \\
			\textbf{Myanmar} & - & - & - & - & 3.9M & 0.9M & - & - & 0.6M \\
			\textbf{Russia} & 13.87 & - & - & 1.1M & - & - & - & - & - \\
			\textbf{Japan} & - & - & - & - & - & - & - & 0.2M & - \\
			\textbf{South Korea} & - & - & - & - & - & - & - & 0.1M & - \\
			\textbf{Sudan} & - & - & - & - & - & - & - & - & 1.1M \\ \hline
			\textbf{\begin{tabular}[c]{@{}l@{}}Share of Global \\ Production (\%)\end{tabular}} & 67.1 & 83.3 & 39.7 & 73.6 & 48.7 & 87.5 & 62.9 & 65.7 & 58.3 \\ \hline
		\end{tabular}%
	}
\end{table*}

\textbf{Deep Weeds:} The Deep Weeds \cite{olsen2019deepweeds} dataset consists of 17,509 low-resolution images of herbaceous rangeland weeds from 9 species. This dataset features a minimum of 1009 images and a maximum of 9016 images per category.

\textbf{IP102:} Wu et al. \cite{wu2019ip102} developed the IP102 dataset to further insect pest recognition research in computer vision. They initially gathered over 300,000 images from popular search engines, which were then labeled by volunteers to ensure relevance to insect pests. Following a data cleaning process, the IP102 dataset consisted of about 75,000 images representing 102 species of common crop insect pests. The dataset also captures various growth stages of some insect pest species.

\textbf{PDD271:} Liu et al. \cite{liu2021plant} developed a large-scale dataset to support plant disease recognition research, consisting of 220,592 images across 271 disease categories. The data was collected in real farm fields with a camera-to-plant distance of 20-30 cm to ensure consistent visual scope. The dataset consists of a minimum of 400 images per category and a maximum of 2000 images.

\begin{figure}
	\centering
	\includegraphics[width=2.5in]{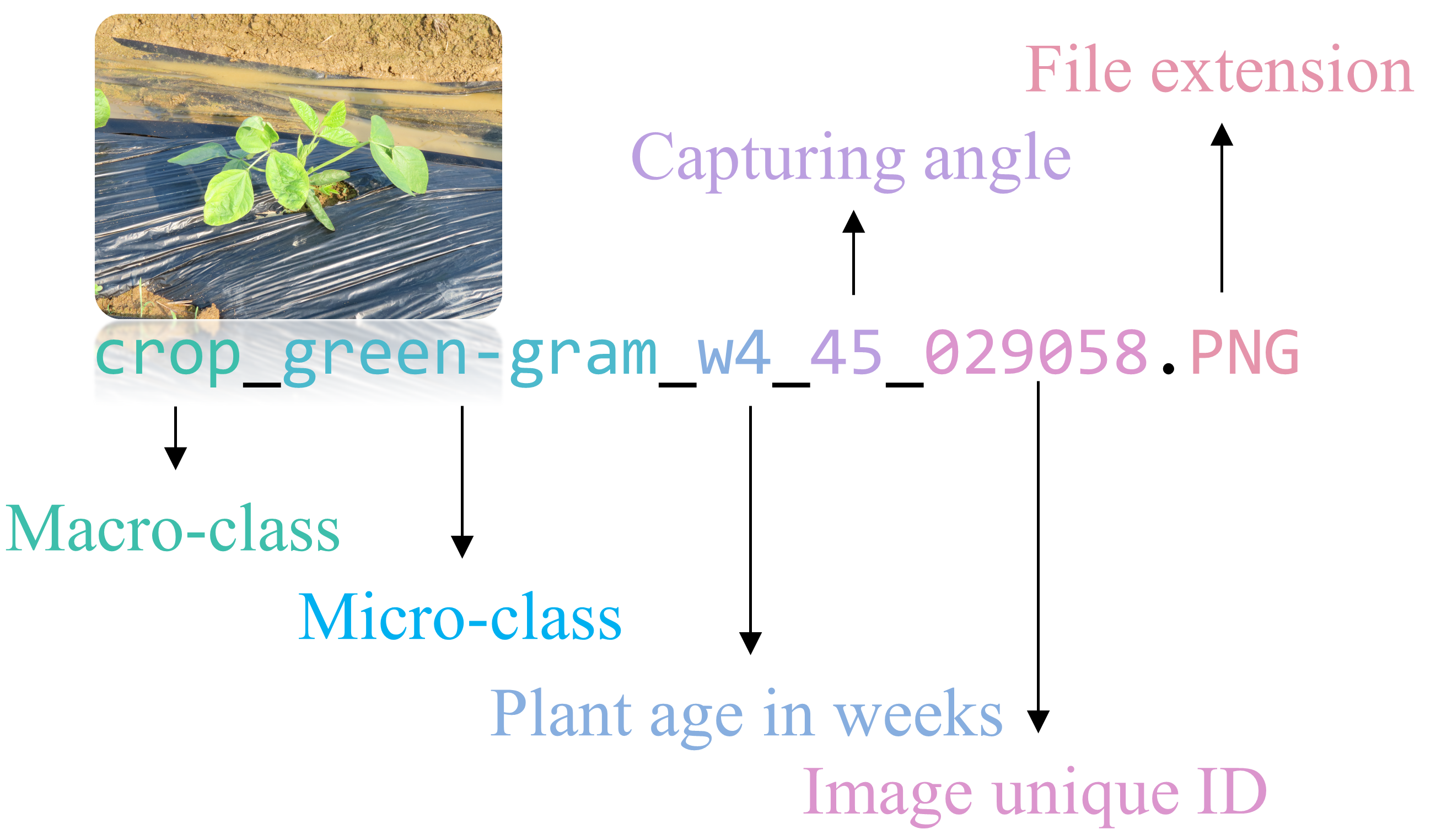}
	\caption{Schematic representation of the file naming convention in the CWD30 dataset, with each segment separated by “\_” indicating specific information about the image, such as species, growth stage, camera angle, and unique ID.}
	\label{fig6}
\end{figure}

Researchers are actively working on plant recognition, frequently utilizing image databases containing samples of particular species to evaluate their methods. The creation of a database necessitates significant time, planning, and manual labor \cite{giselsson2017public,lu2020survey}. Data is usually captured using an array of equipment, from readily available commercial cameras to custom-built sensors designed for specific data acquisition tasks \cite{coudron2021data,coudron2023usefulness}. Consequently, data collected by various researchers differ in quality, sensor type, and quantity, as well as encompassed distinct species. This leads to a diverse and occasionally sparse dataset, often tailored for highly specialized research.

Compared to previous datasets, our proposed CWD30 dataset is unique in that it not only includes images captured from multiple angles, at various growth stages of the plant under varying weather conditions, but also features full plant images rather than just parts of plants (like leaves or branches) see Figure \ref{fig3}. This allows deep learning models to learn more robust and holistic features for better recognition, differentiation, and feature extraction. By addressing the domain-specific challenges of real-field agricultural environments and providing a diverse, varied, and extensive collection of images, CWD30 not only advances research in the field, but also enhances data efficiency and performance in a wide range of downstream agricultural tasks.Table \ref{tab1} presents the statistical information for various agriculture-related datasets.

\begin{figure}
	\centering
	\includegraphics[width=2.5in]{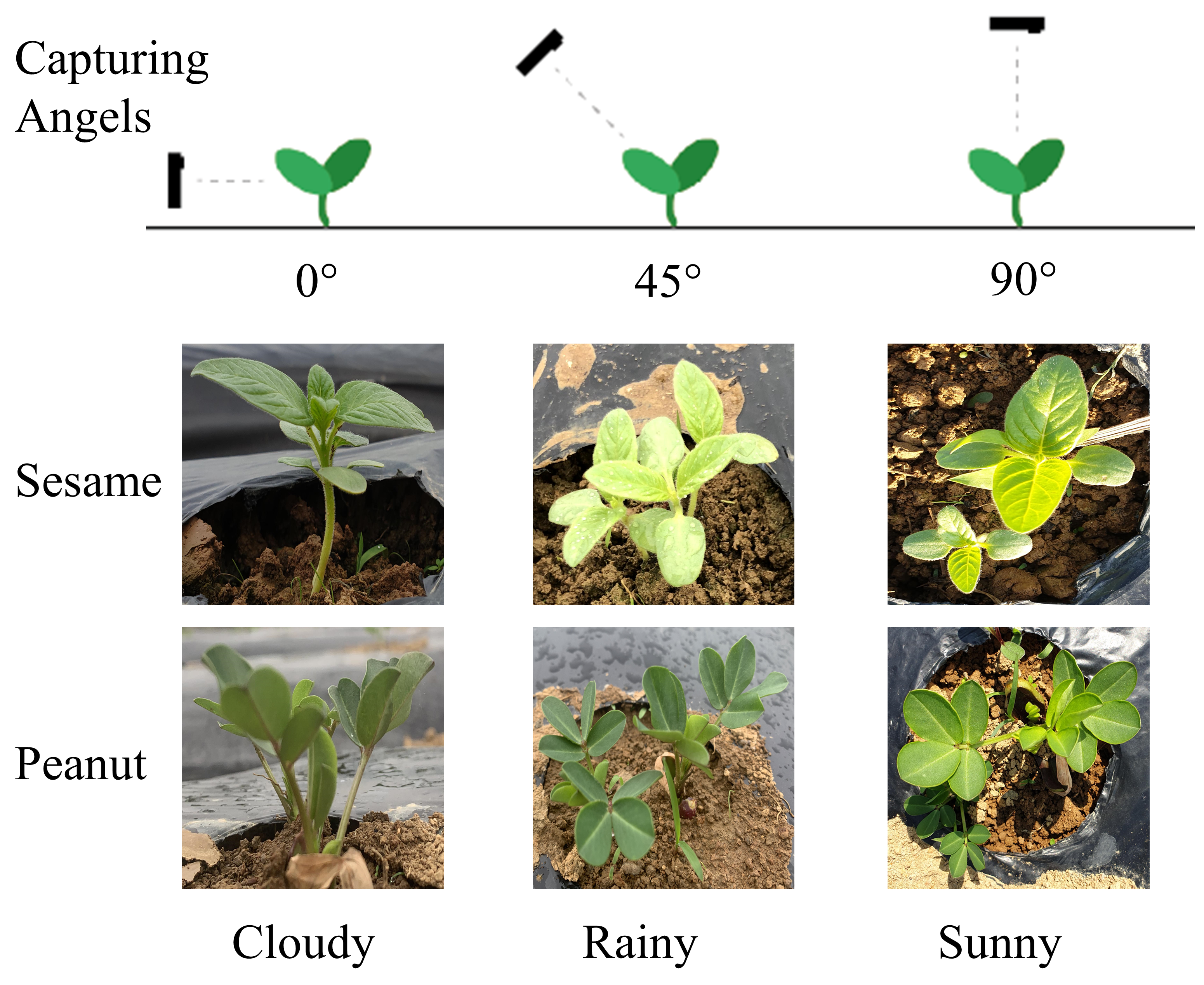}
	\caption{Illustration of camera placement for capturing images at various angles, along with sample images captured at those angles under different weather conditions.}
	\label{fig5}
\end{figure}

%% file: data.tex
\begin{figure*}[!t]
	\centering
	\subfloat[]{\includegraphics[width=4.5in]{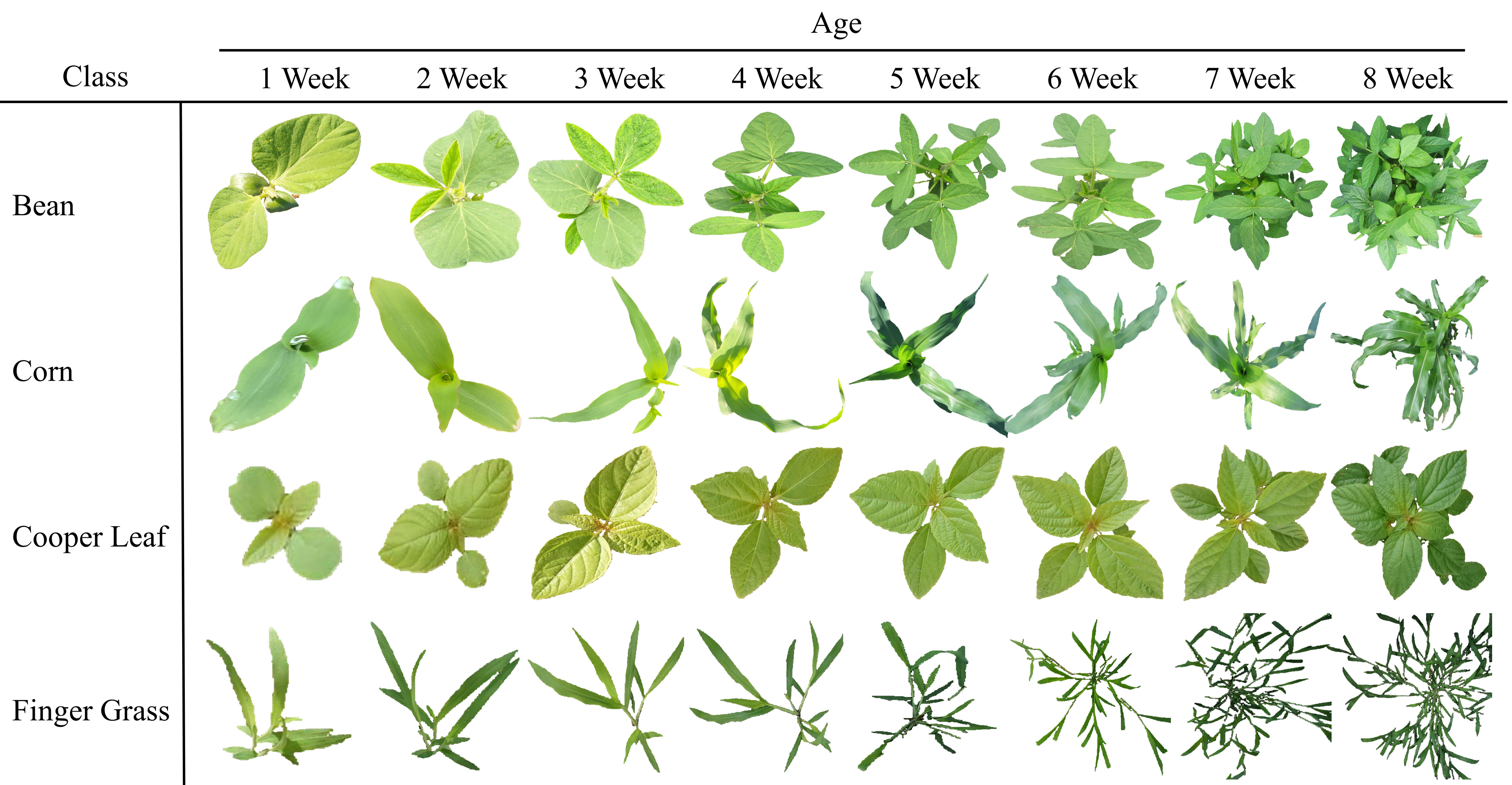}%
		\label{fig7}}
	\hfil
	\subfloat[]{\includegraphics[width=2.5in]{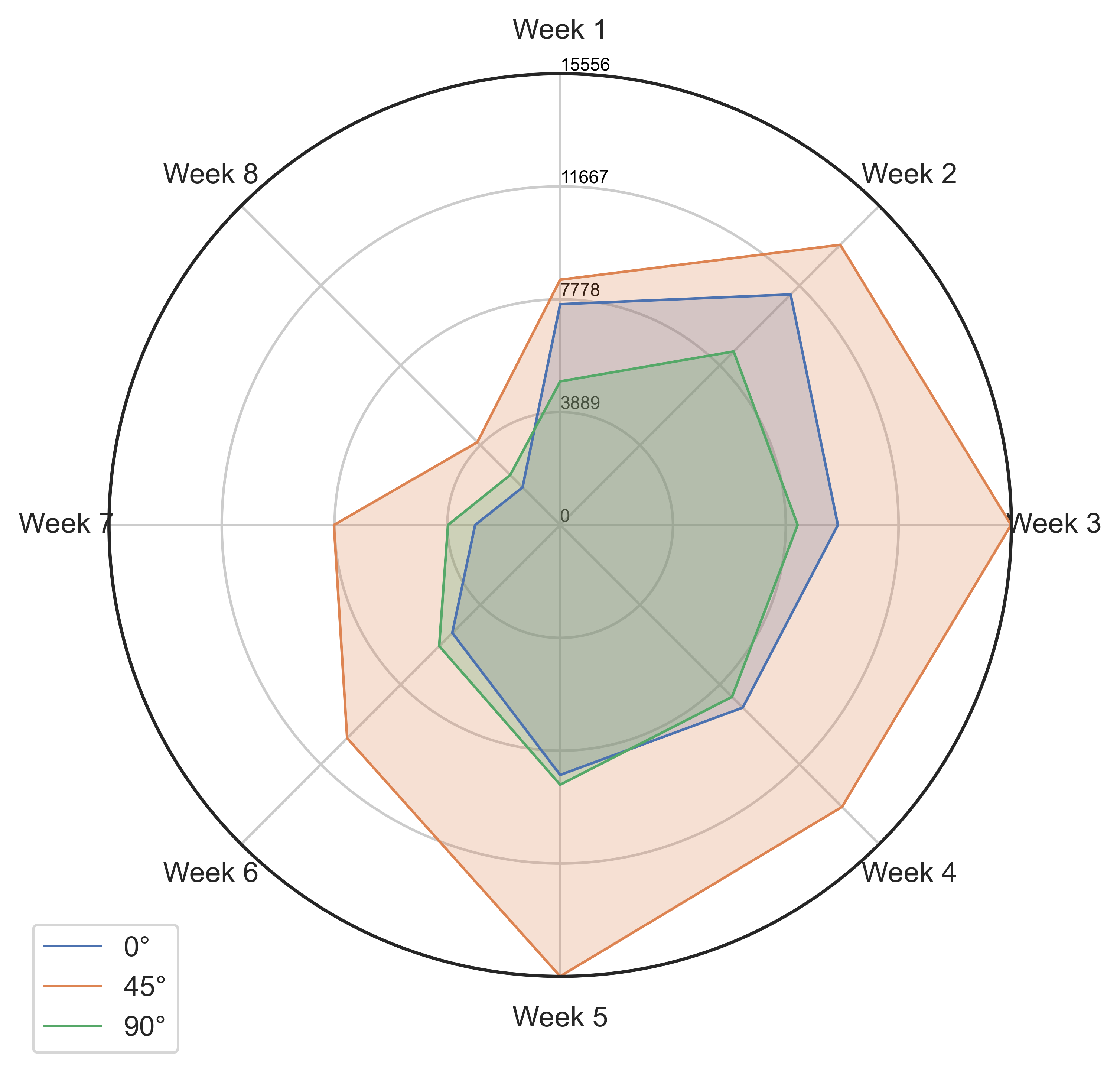}%
		\label{fig8}}
	\caption{(a) A visual representation of plant growth stages, spanning an 8-week period from seedling to maturity, showcasing the developmental progression, changes in color, shape and texture of the plant over time. (b)Radar graph illustrating the distribution of images in the CWD30 dataset across each growing stage during the 8-week period.}
	\label{fig_sim}
\end{figure*}

\section{Data Collection, Preprocessing and Properties}
In this section, we provide a detailed explanation of the collection process, preprocessing, and properties of the proposed CWD30 dataset.
\subsection{Taxonomic System Establishment}
We developed a hierarchical taxonomic system for the CWD30 dataset in collaboration with several agricultural experts from the Rural Development Authority (RDA) in the Republic of Korea. We discussed the most common weed species that affect economically significant crops globally \cite{USDAPlants,WSSA}. A summary of these weeds, the crops they impact, and the countries where they are prevalent is provided in Table \ref{tab2}. We ultimately chose to collect data on approximately 20 of the most problematic weed species worldwide. 
The selection of the 10 crops included in the CWD30 dataset was based on their share in global production and regional importance \cite{RDAKorea,USDAProduction,leff2004geographic}, ensuring the dataset's relevance and applicability in real-world precision agriculture scenarios. Table \ref{tab3} indicates that these crops have considerable shares of global production, with percentages ranging from 39.7\% to 87.5\%. By incorporating crops with substantial importance across various countries, the CWD30 dataset establishes a taxonomy system that addresses the needs of diverse agricultural environments and promotes research in crop recognition and management. 

For weed species not native to Korea, the RDA cultivated them in pots within their facility, as shown in Figure \ref{fig1} (dashed black borders). As for the selected crops, they were divided into two subcategories based on their primary commercial value: economic crops (EC) and field crops (FC). Field crops include staples such as corn and millet, while economic crops encompass legumes (e.g., beans) and oilseeds (e.g., sesame). The resulting hierarchical structure is illustrated in Figure \ref{fig4}. Each crop is assigned both a micro and macro-class based on its properties, whereas weeds are only assigned a micro-class, such as grasses, broad leaves, or sedges. In the CWD30 dataset, we also include a hold-out test set consisting of 23,502 mixed crop and weed (MCW) images, captured both indoors and outdoors, to facilitate the validation of developed models, see Figure \ref{fig2}. We have included a comprehensive table in the appendix of this paper, providing a detailed taxonomy for each plant species within the CWD30 dataset which explains the hierarchical classification, right from the domain, kingdom, and phylum, down to the order, family, genus, and species of each plant.

\begin{figure}[!b]
	\centering
	\includegraphics[width=3in]{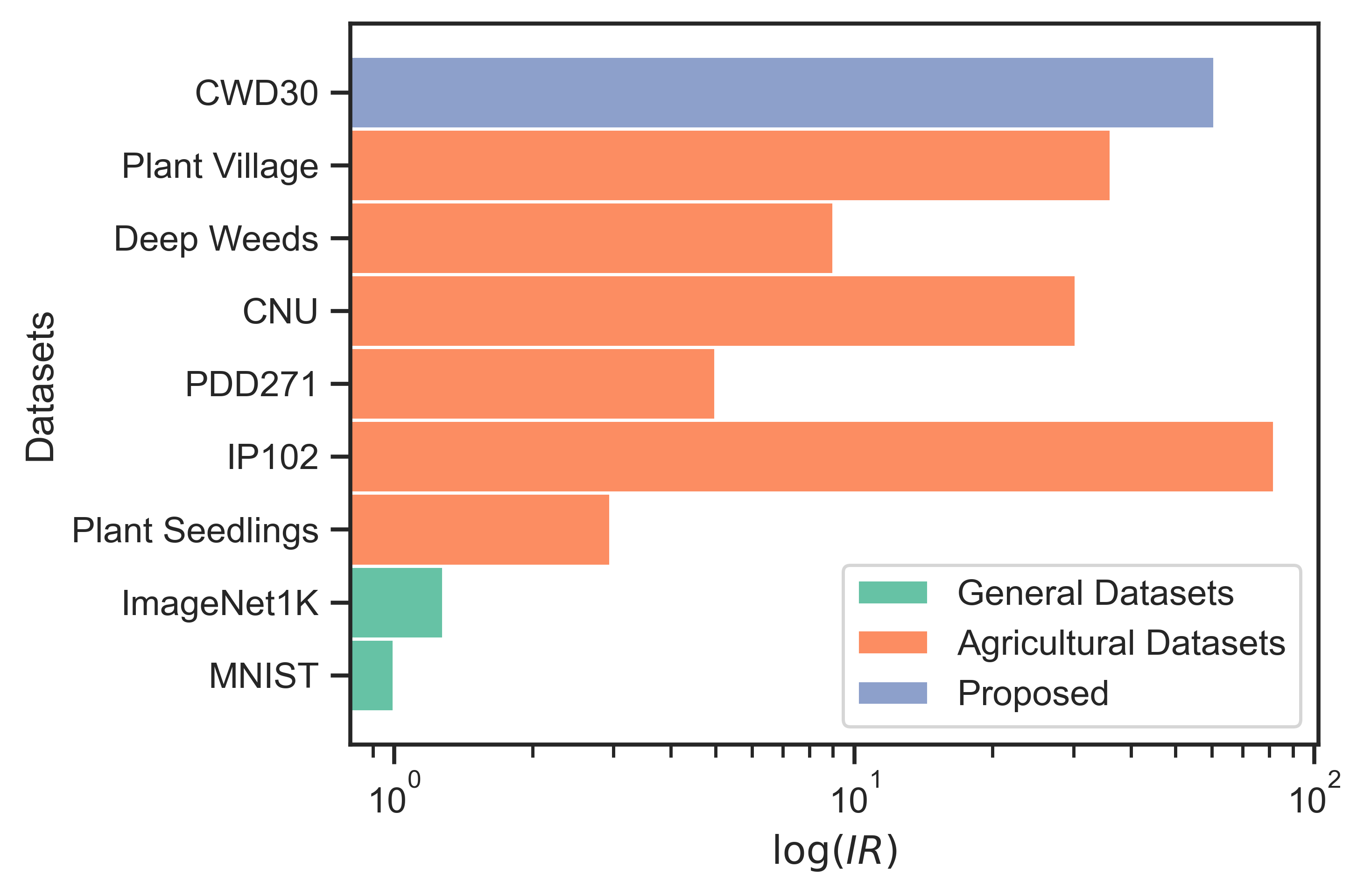}
	\caption{Data imbalance ratio (IR) of proposed dataset in comparison with other datasets.}
	\label{fig9}
\end{figure}

\subsection{Data Collection}
To assemble a benchmark database, we formed five teams: four dedicated to collecting images in farms and fields, and one focused on gathering images from RDA's research facility. Each team was composed of three students from our institute and one field expert. The image collection devices provided to each team varied, including three Canon-SX740 HS, three Canon EOS-200D, three Android phone-based cameras, three iPhone-based cameras, and one DJI Mavic Pro 2.

Each team was tasked with capturing images of two crops and four weeds twice a week. The full dataset is collected over a span of three years from 2020 to 2022. Since image collection is a manual process, the data recorded by different team members varied in quality, perspective, height, sensor type, and species. To ensure diverse data collection, we shuffled the teams monthly and assigned them to collect images of different crops and weeds. This approach helped us obtain a diverse dataset that covers a wide spectrum of real-world challenges and domain shifts, stemming from different sensor types, field environments, and fields of view. Figure \ref{fig1} shows samples of the collected images.

\begin{figure*}[!t]
	\centering
	\includegraphics[width=\textwidth]{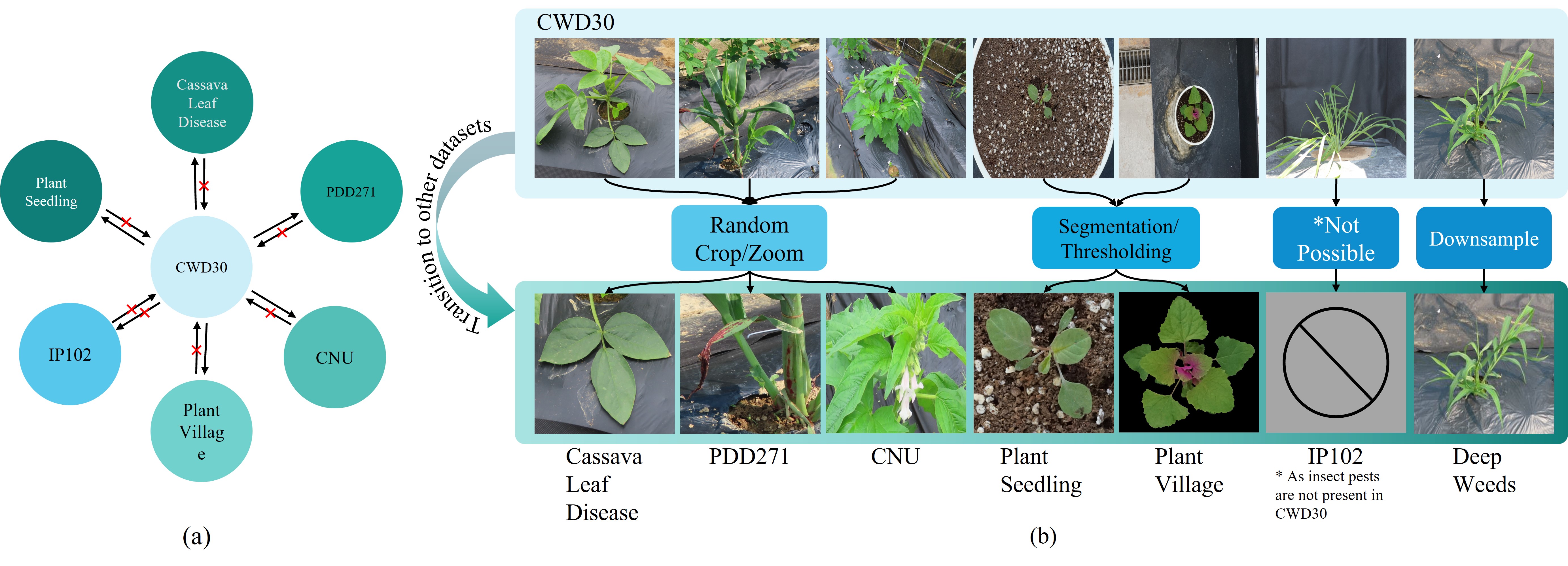}
	\caption{Illustration how simple image processing techniques can transform CWD30 dataset into related subsets, emphasizing CWD30 as a comprehensive superset.}
	\label{fig10}
\end{figure*}

\subsection{Data Filtering, Labelling and Distribution}
The entire data construction process spanned three years. Alongside image collection, five experts reviewed each image to ensure label accuracy monthly. They then removed blurry and noisy images to maintain a clean dataset. The resulting CWD30 dataset comprises 219,778 images, 10 crop types, and 20 weed species. The distribution of each species is depicted in Figure \ref{fig2}. The minimum number of images per species is 210, while the maximum is 12,782. This unbalanced distribution reflects real-world scenarios where it is challenging to obtain data samples for certain classes. In our case, this occurred for weed species that were difficult to cultivate in Korea's weather conditions. As for labeling, each file is saved with a unique naming format, an example of which can be seen in Figure \ref{fig6}.

\subsection{Data Splits}
The CWD30 dataset comprises 219,778 images and 30 plant species. To ensure more reliable test results, we employed a K-fold  validation method with K=3, guaranteeing enough samples for each category in the testing set \cite{yadav2016analysis}. We divided the data into three randomized folds for training (74,724), validation (72,526), and testing (72,526), adhering to a 0.33:0.33:0.34 split ratio. For each fold, we partitioned every plant species into three sections, taking care to include an equal proportion of the smallest class within each section (refer to Figure \ref{fig2}). The training, validation, and testing sets were split at the micro-class level.

\begin{table}[!b]
	\centering
	\caption{Classification performance of various deep learning models on the CWD30 dataset, comparing results obtained from random initialization and ImageNet initialization.}
	\label{tab04}
	\renewcommand{\arraystretch}{1.25}
	\resizebox{3.5in}{!}{%
		\begin{tabular}{lllllllllll}
			\cline{1-6}
			\multicolumn{1}{c}{\multirow{2}{*}{\textbf{Typ.}}} & \multicolumn{1}{c}{\multirow{2}{*}{\textbf{Methods}}} & \multicolumn{2}{c}{\textbf{Scratch}} & \multicolumn{2}{c}{\textbf{ImageNet-1K}} &  \\ \cline{3-6}
			\multicolumn{1}{c}{} & \multicolumn{1}{c}{} & \multicolumn{1}{c}{\textbf{F1}} & \multicolumn{1}{c}{\textbf{Acc}} & \multicolumn{1}{c}{\textbf{F1}} & \multicolumn{1}{c}{\textbf{Acc}} &  \\ \cline{1-6}
			\multicolumn{1}{c}{\multirow{4}{*}{CNN}} & ResNet-101\cite{he2016deep} & 76.38 & 80.17 & 83.83 & 88.66 &  \\
			\multicolumn{1}{c}{} & ResNext-101\cite{xie2017aggregated} & 79.76 & 81.36 & 84.03 & 89.06 &  \\
			\multicolumn{1}{c}{} & MobileNetv3-L\cite{howard2019searching} & 74.67 & 78.95 & 81.80 & 86.29 &  \\
			\multicolumn{1}{c}{} & EfficientNetv2-M\cite{tan2021efficientnetv2} & 87.37 & 83.06 & 84.91 & 90.79 &  \\ \cline{1-6}
			\multicolumn{1}{c}{\multirow{3}{*}{Trans.}} & ViT\cite{dosovitskiy2020image} & 78.90 & 83.43 & 84.08 & 87.84 &  \\
			\multicolumn{1}{c}{} & SwinViT\cite{liu2021swin} & 81.53 & 87.59 & 83.70 & 88.71 &  \\
			\multicolumn{1}{c}{} & MaxViT\cite{tu2022maxvit} & 82.24 & 87.08 & 82.43 & 91.45 &  \\ \cline{1-6}
		\end{tabular}%
	}
\end{table}

\subsection{Viewing Angles and Growth Stages}
Our proposed CWD30 dataset stands out from previous datasets due to its unique and beneficial properties, with three prominent features: (i) images captured from multiple angles, (ii) images taken at various growth stages and under varying weather conditions, and (iii) full plant images instead of just plant parts like leaves or branches. These characteristics enable deep learning models to learn more robust and comprehensive features for enhanced recognition, differentiation, and feature extraction.

Capturing plant images from different angles for deep learning models results in robust feature learning, improved occlusion handling, scale and rotation invariance, and better management of lighting and shadow variations. This leads to more accurate and reliable CAPA systems that perform well in real-world agricultural environments. Figure \ref{fig5} depicts a visual representation of the various angles used for image collection.

Furthermore, the growing interest in plant phenomics and the use of image-based digital phenotyping systems to measure morphological traits has led to increased efforts to bridge the genotyping-phenotyping gap. However, research in this area is limited, mainly due to the lack of available datasets providing whole-plant level information rather than specific plant parts, such as leaves, nodes, stems, and branches. The CWD30 dataset, which includes full plant images from multiple viewing angles and at different growth stages, can accelerate our understanding of genotype-phenotype relationships. It can also assist plant scientists and breeders in developing advanced phenotyping systems that offer more detailed phenotypic information about plants. Figure \ref{fig7} displays randomly selected samples of crops and weeds at different life cycle stages, with images captured at a 90-degree angle from the plant. The graph in Figure \ref{fig8} show the distribution of images across each growing stage.

\begin{figure}[!b]
	\centering
	\includegraphics[width=3in]{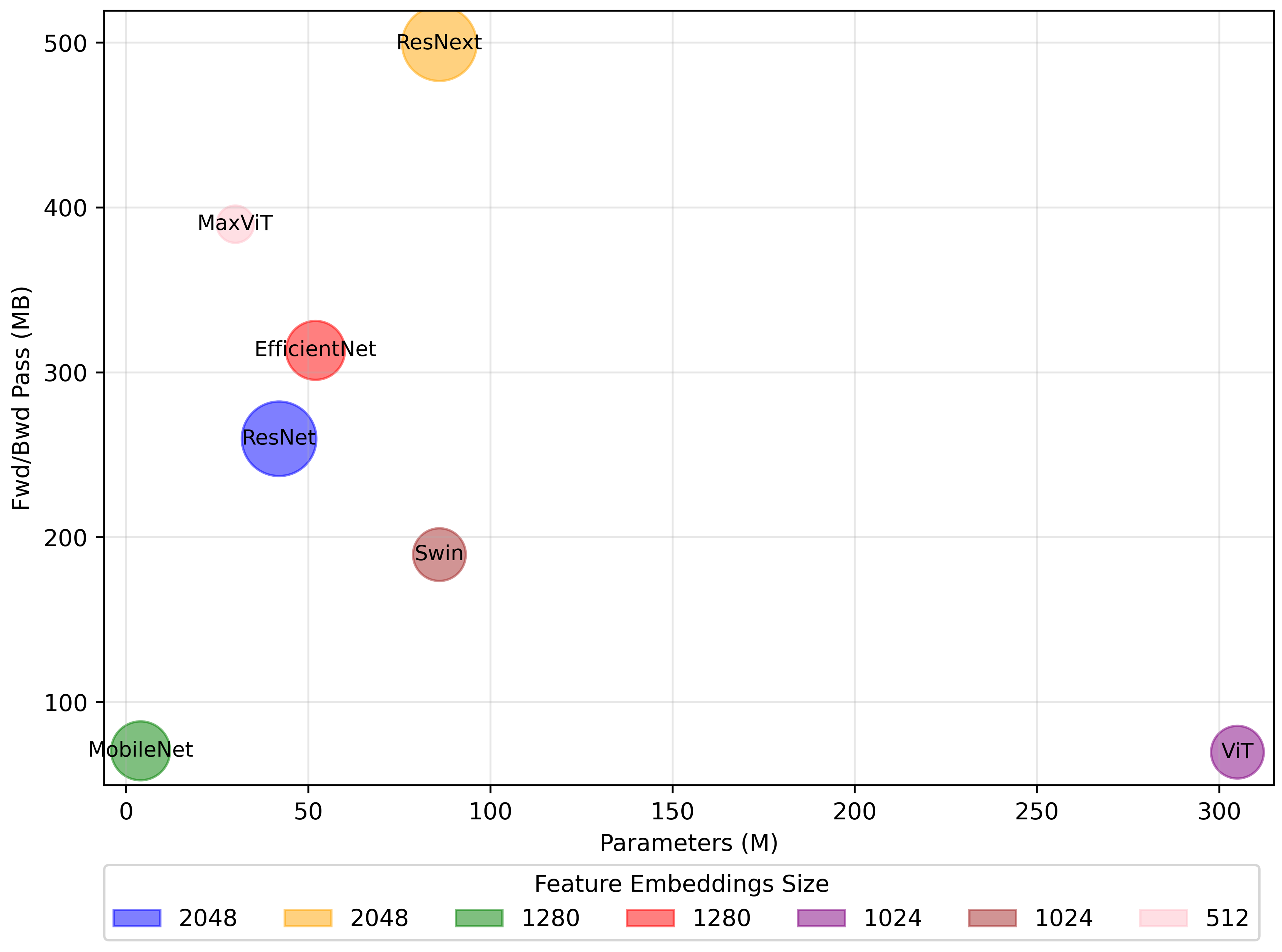}
	\caption{Graph comparing deep learning models in terms of parameters (in million), feature embeddings (no. of features), and forward and backward pass sizes (in megabytes), highlighting the trade-offs among the models. Best viewed in color.}
	\label{fig11}
\end{figure}

\subsection{Comparison with Other Datasets}
In this section, we compare the CWD30 dataset with several existing datasets related to crop-weed recognition. Our dataset stands out as a more holistic, domain-adverse, versatile, and diverse dataset that provides a comprehensive solution to crop-weed discrimination. Furthermore, it classifies weeds into major families, such as grasses, sedges, and broad leaves, and further into specific weed sub-categories. To the best of our knowledge, CWD30 is the first dataset of its kind in the field of practical crop-weed discrimination.

The PDD271 dataset contains close-up images of only diseased plant parts, the Deep Weeds dataset has low-resolution images of roadside weeds, and the Plant Seedling dataset consists of early-stage weeds grown in lab trays. The most comparable dataset in this field is the CNU weed dataset, which focuses on field environments but features simplified representations of plants, i.e., zoomed in part of plants.

Existing data sets’ shortcomings can be summarized as follows:
\begin{enumerate}
\item Simplified representation: By focusing on specific plant parts, such as leaves or branches, the data becomes less complex and fails to represent real-field challenges.
\item Limited scope: Images of specific plant parts may not capture the full characteristics of a plant, leading to less accurate recognition systems.
\item Restricted environments: Capturing images in specific fields may limit the model's ability to generalize to other settings or conditions.
\item Less robust features: The absence of multiple angles and growth stages may result in less robust feature learning and hinder the model's ability to handle occlusions, rotations, and scale variations.
\item Smaller dataset size: Most existing precision agricultural datasets have a limited number of images, hindering the development of more advanced deep learning-based systems.
\end{enumerate}
In contrast, the CWD30 dataset addresses these limitations with the following inherent properties:
\begin{enumerate}
\item Comprehensive representation: Full-plant images provide a holistic view, capturing various aspects of the crops and weeds.
\item Varied environments: Capturing plants in both indoor and outdoor settings enable the dataset to cover a broader range of conditions and will enhance the model's generalizability.
\item Multiple angles: Images taken from different angles allow models to learn robust features and improve occlusion handling, rotation invariance, and scale invariance.
\item Different growth stages: Capturing images at various growth stages helps models recognize crops and weeds at any stage of their life cycle, resulting in more accurate and reliable CAPA systems.
\item Complexity: Increased variability and complexity make the images more challenging to analyze.
\item Larger dataset size: The proposed dataset is one of the largest real-image datasets to date in the field of precision agriculture.
\end{enumerate}

By addressing domain-specific challenges in real-field agricultural environments and providing a diverse, varied, and extensive collection of images, CWD30 advances research in the field and enhances data efficiency and performance in a wide range of downstream agricultural tasks.

An additional advantage of the CWD30 dataset is its versatility, which allows it to encompass various existing agricultural datasets through simple image processing operations. By applying random cropping, downsampling, foreground segmentation, or thresholding to the images in the CWD30 dataset, one can create subsets that resemble other datasets in the field. An example of this process is shown in Figure \ref{fig9}. This demonstrates that the CWD30 dataset can be considered a comprehensive and unified source of agricultural data, with other datasets effectively serving as subsets of CWD30. This versatility not only highlights the extensive nature of the CWD30 dataset but also supports its potential for advancing research and improving performance in a wide range of agricultural tasks.

\begin{table*}[!t]
	\caption{Performance comparison of deep learning models using pretrained weights from ImageNet and CWD30, highlighting the impact of dataset-specific pretraining on model performance.}
	\label{tab5}
	\renewcommand{\arraystretch}{1.2}
	\resizebox{\textwidth}{!}{%
	\begin{tabular}{llllllllll}
		\hline
		\multicolumn{1}{c}{\multirow{2}{*}{\textbf{Typ.}}} & \multicolumn{1}{c}{\multirow{2}{*}{\textbf{Methods}}} & \multicolumn{2}{c}{\textbf{Deep Weeds}\cite{olsen2019deepweeds}} & \multicolumn{2}{c}{\textbf{Plant Seedlings}\cite{giselsson2017public}} & \multicolumn{2}{c}{\textbf{Cassava Plant}\cite{ayu2021deep}} & \multicolumn{2}{c}{\textbf{IP 102}\cite{wu2019ip102}} \\ \cline{3-10} 
		\multicolumn{1}{c}{} & \multicolumn{1}{c}{} & \multicolumn{1}{c}{\textbf{ImageNet-1k}} & \multicolumn{1}{c}{\textbf{CWD-30}} & \multicolumn{1}{c}{\textbf{ImageNet-1k}} & \multicolumn{1}{c}{\textbf{CWD-30}} & \multicolumn{1}{c}{\textbf{ImageNet-1k}} & \multicolumn{1}{c}{\textbf{CWD-30}} & \multicolumn{1}{c}{\textbf{ImageNet-1k}} & \multicolumn{1}{c}{\textbf{CWD-30}} \\ \hline
		\multirow{4}{*}{CNN} & ResNet-101\cite{he2016deep} & 91.13 & 95.08 & 90.14 & 96.27 & 64.82 & 71.44 & 60.34 & 66.87 \\
		& ResNext-101\cite{xie2017aggregated} & 90.70 & 95.87 & 92.46 & 97.79 & 65.01 & 73.22 & 62.13 & 67.90 \\
		& MobileNetv3-L\cite{howard2019searching} & 89.08 & 94.62 & 88.43 & 96.54 & 66.34 & 71.17 & 61.08 & 64.53 \\
		& EfficientNetv2-M\cite{tan2021efficientnetv2} & 91.39 & 95.78 & 90.85 & 97.18 & 61.13 & 69.34 & 60.86 & 68.29 \\ \hline
		\multirow{3}{*}{Trans.} & ViT\cite{dosovitskiy2020image} & 86.25 & 90.18 & 91.41 & 95.39 & 58.24 & 61.32 & 59.77 & 68.46 \\
		& SwinViT\cite{liu2021swin} & 88.83 & 96.70 & 93.24 & 98.06 & 73.83 & 78.66 & 59.11 & 68.67 \\
		& MaxViT\cite{tu2022maxvit} & 87.79 & 97.04 & 92.47 & 97.89 & 71.55 & 79.54 & 60.51 & 69.36 \\ \hline
	\end{tabular}%
	}
\end{table*}

\subsection{Data Imbalance Ration}
A dataset's imbalance ratio (IR) refers to the degree of disparity between the number of samples in different classes \cite{johnson2019survey}. In the context of deep learning, the imbalance ratio can have significant effects on model performance. Although low data imbalance ratios in datasets, like MNIST and ImageNet-1K, are generally preferred for deep learning models as they promote balanced class representation and accurate performance, these datasets do not always represent real-world situations where data samples for some classes are harder to obtain.

In contrast, high data imbalance ratios, found in datasets such as CNU, CWD30, and DeepWeeds, can pose challenges for deep learning models as they may lead to overfitting and poor generalization. Models trained on highly imbalanced datasets can become biased towards majority classes, resulting in decreased performance for minority classes. However, one key advantage of having high imbalance ratios is their increased representation of real-world situations, particularly in complex recognition tasks like precision agriculture, where some classes naturally have fewer available samples. While these imbalanced datasets present challenges, they also offer a more realistic depiction of real-world scenarios, pushing deep learning models to adapt and improve their performance in diverse and unevenly distributed data conditions. Figure \ref{fig10} shows imbalance ration of related datasets.

To the best of our knowledge, the proposed CWD30 dataset offers several distinctive features not found in previous datasets, as highlighted in earlier sub-sections. These features can bridge the genotyping-phenotyping gap, enhance the robustness and reliability of deep learning systems, and expand their area of applications.

%% file: experiments.tex
\section{Experiments and Evaluation}
We conducted a comprehensive experimental evaluation of the CWD30 dataset, focusing on classification performance using deep convolutional and transformer-based architectures. Additionally, we examine the influence of CWD30 pretrained networks on downstream precision agriculture tasks, including semantic segmentation.

\subsection{Experimental Setup}
In our experiments all networks' layers are fine-tuned using an AdamW optimizer with a minibatch size of 32 and an initial learning rate of 6e-5. We employ a cosine decay policy for reducing the learning rate and incorporate a dropout value of 0.2, along with basic data augmentations, to prevent overfitting. While the deep models' fundamental architectures remain unchanged, the last fully connected layer is adapted to match the number of target classification classes. Each network is trained for 50 epochs across all datasets, and the reported results represent the average of three runs. Input images are resized to 224 x 224 pixels. Our deep feature-based experiments are implemented using PyTorch and performed on an NVIDIA Titan RTX-3090 GPU with 24 GB of onboard memory.

\subsection{Evaluation Metrics}
To objectively assess models trained on the CWD30 dataset, we employ widely accepted evaluation metrics for comprehensive comparisons. Given the dataset's imbalanced class distribution, we utilize the following metrics for better performance assessment:
\begin{itemize}
\item	Per-class Mean Accuracy (Acc) calculates the average of individual class mean accuracies, providing a balanced performance evaluation, especially for imbalanced datasets like CWD30.
\item	F1-Score is the harmonic mean of precision (the ratio of true positive predictions to the sum of true positive and false positive predictions) and recall (the ratio of true positive predictions to the sum of true positive and false negative predictions), offering a single value representing the model's overall performance while accounting for false positive and false negative errors.
\item For downstream tasks like semantic segmentation, we use mean intersection over union (mIoU), which evaluates the overlap between predicted and ground truth segments.
\end{itemize}
By examining these metrics, researchers can identify the most promising approaches to guide future developments in precision agriculture and the development of CAPA systems.

%% file: results.tex
\begin{figure*}
	\centering
	\includegraphics[width=\textwidth]{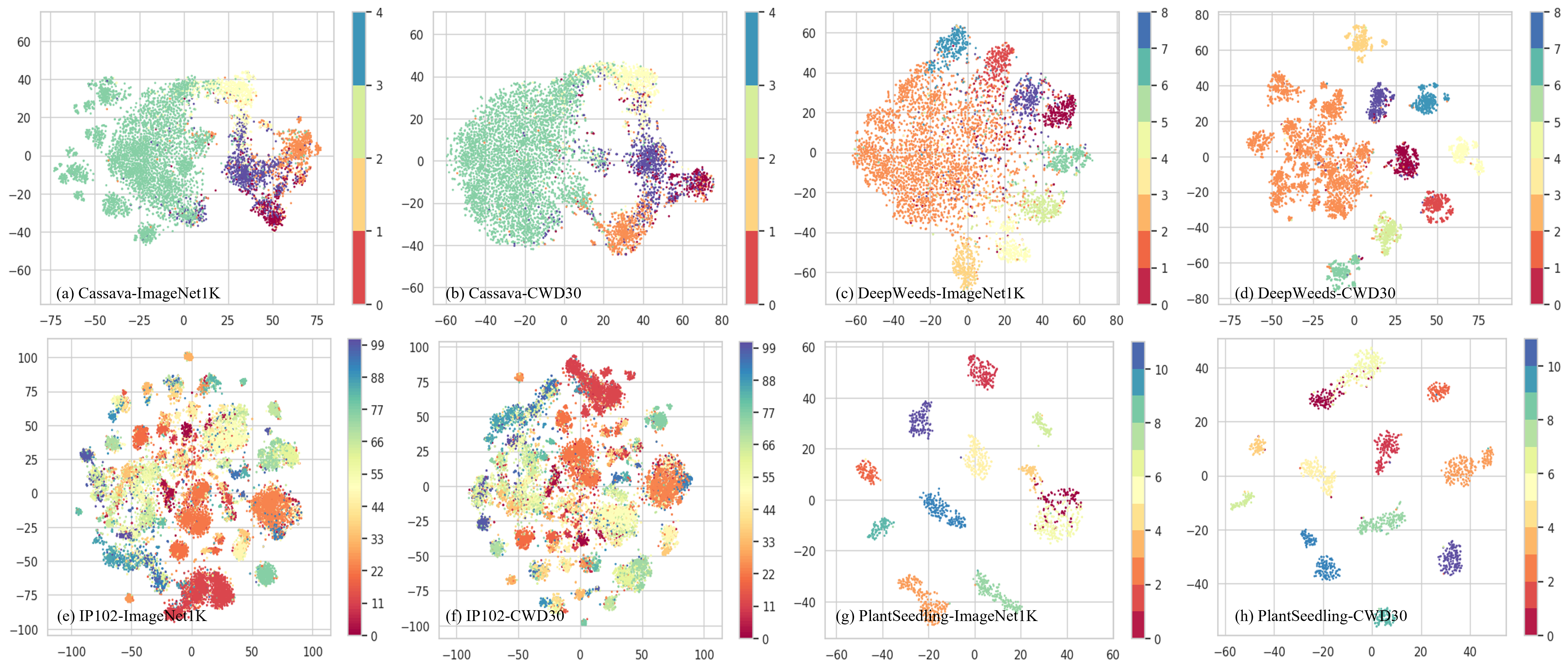}
	\caption{2D t-SNE feature embeddings visualization comparing best performing deep learning model (i.e., MaxViT) with pretrained weights from ImageNet and CWD30, on various agricultural datasets. Highlighting the improved cluster patterns and separation achieved using the CWD30 pretrained network.}
	\label{fig12}
\end{figure*}

\section{Results and Discussion}
In this section, we present the classification results for various deep learning models trained on the CWD30 dataset. We compare the models \cite{tan2021efficientnetv2,liu2021swin,tu2022maxvit,he2016deep,xie2017aggregated,howard2019searching,dosovitskiy2020image} based on their F1-Score and per-class mean accuracy (Acc) when trained from scratch and when pretrained on the ImageNet-1K dataset. The results are summarized in the table \ref{tab04}. The results reveal that EfficientNetv2-M \cite{tan2021efficientnetv2} is the best-performing CNN architecture when trained from scratch, with the highest F1-Score (82.37) and accuracy (87.06). Pretraining on ImageNet-1K consistently improves the performance of all models. Among transformer-based models, SwinViT \cite{liu2021swin} achieves the highest accuracy (88.71), and MaxViT  \cite{tu2022maxvit} obtains the highest F1-Score (82.43). Generally, more complex models like EfficientNetv2-M and MaxViT outperform less complex counterparts, as their increased capacity better captures and represents the nuances in the CWD30 dataset.

Moreover, transformer-based models like SwinViT and MaxViT demonstrate superior performance compared to their CNN counterparts despite having fewer parameters and a smaller memory footprint (forward and backward pass). This observation underscores the potential of transformer architectures for handling the diverse and complex patterns in the CWD30 dataset. The self-attention mechanism in transformers may allow them to capture long-range dependencies and fine-grained patterns more effectively than traditional convolutional layers.

Additionally, we compare the model parameters and memory footprint against the final output feature embeddings generated by the model just before the linear classification layers, as shown in the figure \ref{fig10}. Intriguingly, MaxViT, which outputs the fewest feature embeddings (512), still outperforms all other models. This finding is significant because lower-dimensional feature embeddings offer practical advantages for real-world applications, especially in resource-constrained environments. For instance, in precision agriculture, heavy GPUs like the RTX-3090 may not be suitable for field deployment due to their large size and power consumption. Instead, smaller embedded systems like NVIDIA Jetson boards are commonly used, which have limited memory and computational resources. By employing deep learning models with lower-dimensional embeddings, parameters, and memory footprint, these systems can efficiently process and analyze data, making them more suitable for real-world applications.

The diverse and sizable CWD30 dataset is essential for the development of robust and reliable CAPA systems, as it offers a rich source of real-world precision agriculture data for training deep data hungry models. By focusing on the quality of the dataset and addressing practical constraints of real-world deployments, researchers can ensure that deep learning models are capable of handling inherent variability and imbalances in agricultural settings, ultimately making them more efficient, generalizable, and suitable for a wide range of applications, including field deployment.

\begin{figure}
	\centering
	\includegraphics[width=3in]{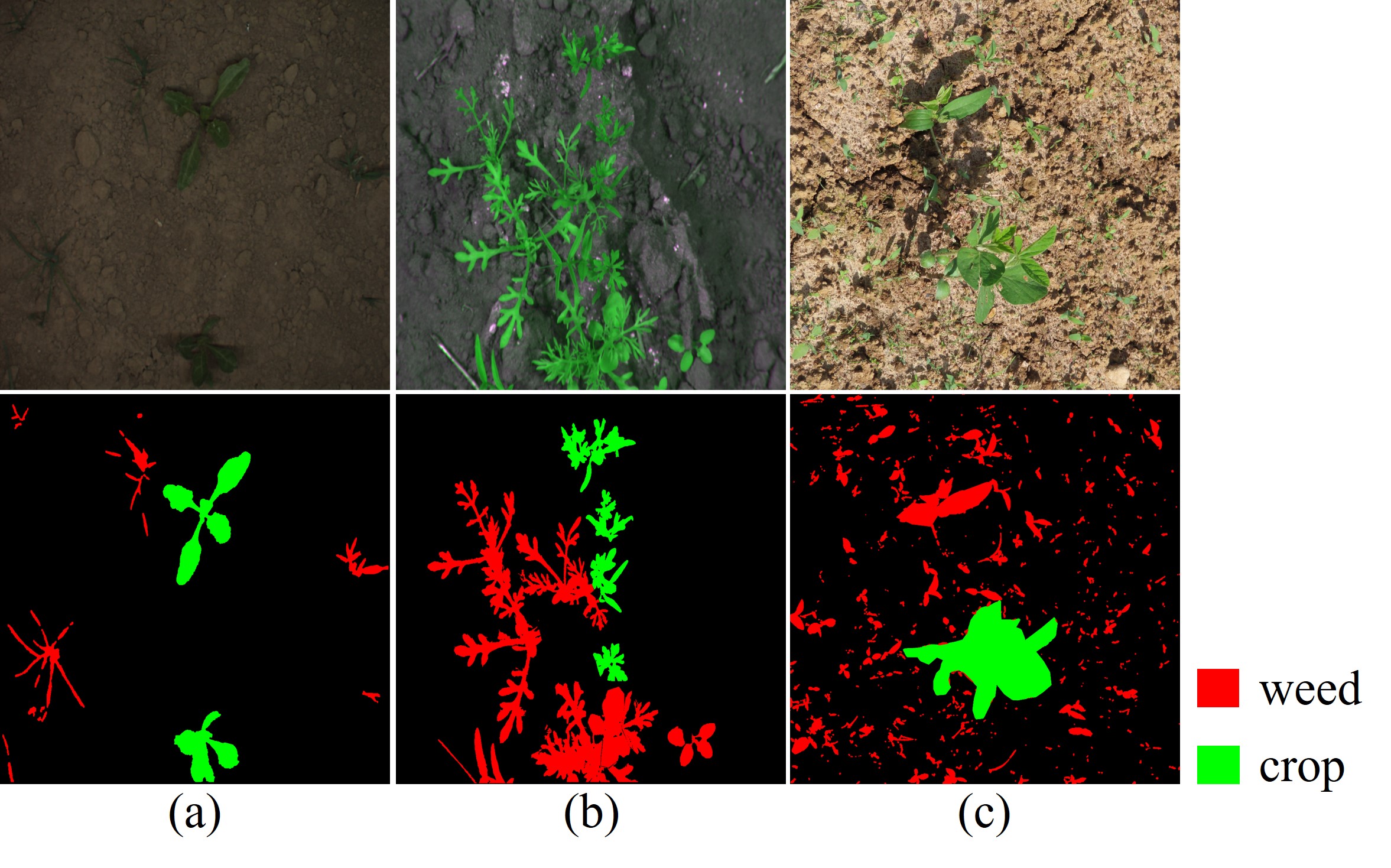}
	\caption{Sample images from (a) SugarBeet \cite{chebrolu2017ijrr}, (b) BeanWeed \cite{ilyas4345158adaptive} and (c) CarrotWeed \cite{haug15} datasets.}
	\label{fig13}
\end{figure}

\subsection{Further Analysis}
To further evaluate the performance enhancements offered by using the CWD30 dataset for pretraining and finetuning on tasks with limited samples, we tested multiple publicly available benchmark agricultural datasets \cite{olsen2019deepweeds,giselsson2017public,wu2019ip102,ayu2021deep} for robust feature extraction and compared the results with models pretrained on the ImageNet-1K dataset. Detailed information about these datasets is provided in Section II. For each dataset, we adhere to the testing and data split settings outlined in their original papers, while maintaining the same network training settings as described in the previous subsection. The results are summarized in Table \ref{tab5}. Throughout all datasets MaxViT achieved highest per class mean accuracy scores despite having minimum output feature embeddings.  Whereas pretraining on CWD30 dataset consistently improves the performance of all tested architectures on all datasets.

For better understanding and comparison, we extract high-dimensional feature embeddings (features of second last layer) from the best-performing model, i.e., MaxViT, on test images of all datasets. The compactness and expressiveness of these feature embeddings facilitate the development of efficient and accurate algorithms for various applications, including CAPA systems. We perform t-SNE \cite{cai2022theoretical} visualization on these feature embeddings. t-SNE, effectively projects high-dimensional feature embeddings onto a two-dimensional space while preserving the local structure and relationships within the data. By plotting t-SNE visualizations, we can assess the separability and distribution of the data in the reduced space, as well as the quality of the learned feature representations.

Our results reveal that models pretrained on the CWD30 dataset produce more distinct and well-separated clusters in the t-SNE plots when fine-tuned on various public datasets compared to ImageNet pretrained models. The t-SNE plots for CWD30 and ImageNet pretrained MaxViT models on publicly available datasets are displayed in Figure \ref{fig12}. From the Figure \ref{fig12}, it is evident that CWD30-pretrained models learn more meaningful and robust feature representations, as the clusters in these plots are better defined and distinct, with points belonging to the same cluster positioned closer together and clear separation between clusters. This ultimately leads to improved performance during finetuning and downstream tasks (see section V.B).

\begin{table*}[!htb]
	\centering
	\caption{Comparison of performance on downstream segmentation tasks using pretrained backbones (i.e., ImageNet vs. CWD30).}
	\label{tab6}
	\renewcommand{\arraystretch}{1.2}
	\resizebox{\textwidth}{!}{%
		\begin{tabular}{lllllllllll}
			\cline{1-8}
			\multicolumn{1}{c}{\multirow{2}{*}{\textbf{Method}}} & \multicolumn{1}{c}{\multirow{2}{*}{\textbf{Backbone}}} & \multicolumn{2}{c}{\textbf{SugarBeet}} & \multicolumn{2}{c}{\textbf{CarrotWeed}} & \multicolumn{2}{c}{\textbf{BeanWeed}} &  \\ \cline{3-8}
			\multicolumn{1}{c}{} & \multicolumn{1}{c}{} & \multicolumn{1}{c}{\textbf{ImageNet-1k}} & \multicolumn{1}{c}{\textbf{CWD-30}} & \multicolumn{1}{c}{\textbf{ImageNet-1k}} & \multicolumn{1}{c}{\textbf{CWD-30}} & \multicolumn{1}{c}{\textbf{ImageNet-1k}} & \multicolumn{1}{c}{\textbf{CWD-30}} &  \\ \cline{1-8}
			U-Net\cite{ronneberger2015u} & ResNet-101\cite{he2016deep} & 80.96 & 85.47 & 75.47 & 78.32 & 69.67 & 72.49 &  \\
			DeepLabv3+\cite{chen2017rethinking} & ResNet-101\cite{he2016deep} & 81.17 & 86.02 & 80.29 & 83.16 & 72.41 & 78.03 &  \\
			OCR\cite{yuan2020object} & ResNet-101\cite{he2016deep} & 84.79 & 87.34 & 84.56 & 86.53 & 73.60 & 79.51 &  \\
			SegNeXt-L\cite{guo2022segnext} & MSCAN\cite{guo2022segnext} & 84.15 & 87.65 & 83.79 & 88.54 & 80.05 & 83.90 &  \\ \cline{1-8}
		\end{tabular}%
	}
\end{table*}

\subsection{Performance on Downstream Tasks}
To evaluate the effectiveness of enhanced feature representations obtained by CWD30 pretraining on downstream tasks, we assess several state-of-the-art segmentation models for pixel-level crop weed recognition. We use three publicly available crop-weed datasets: CarrotWeed \cite{haug15}, SugarBeet \cite{chebrolu2017ijrr}, and BeanWeed \cite{ilyas4345158adaptive}. Sample images from each dataset, along with their corresponding segmentation labels, are shown in Figure. The quantitative results are summarized in Table \ref{tab6}.

\begin{figure*}[!t]
	\centering
	\includegraphics[width=\textwidth]{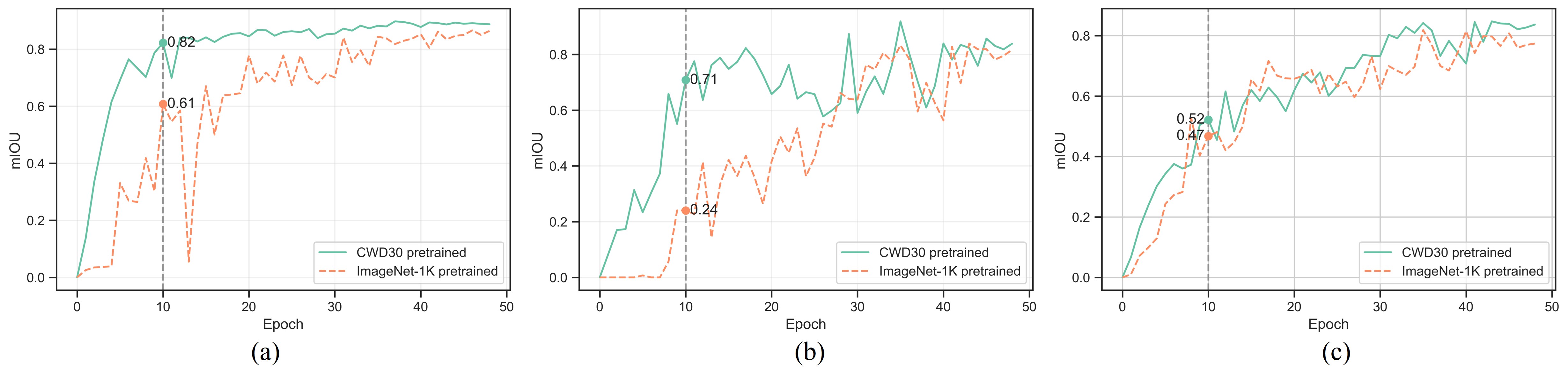}
	\caption{Learning curves illustrating the superior performance and faster convergence of CWD30 pretrained backbones on downstream segmentation tasks.(a) SugarBeet \cite{chebrolu2017ijrr}, (b) CarrotWeed \cite{haug15} and (c) BeanWeed \cite{ilyas4345158adaptive}.}
	\label{fig14}
\end{figure*}

Throughout the experiments, it is evident that pretraining architecture backbones with CWD30 provides a clear advantage over ImageNet-1K pretrained backbones. Although the performance difference may not appear substantial when examining the table \ref {tab6}, the difference becomes more apparent when analyzing the learning curves of both setups. The learning curves of the best-performing SegNext \cite{guo2022segnext} model are shown in Figure \ref{fig14}. These curves demonstrate that initializing experiments with weights obtained from training on more relevant datasets (i.e., agricultural data) results in faster convergence and stable training. From the plots, it can be seen that the difference between ImageNet and CWD30 initialization is significant at the 10th epoch, where the CWD30-initialized model already reaches performance close to its final convergence value. In contrast, for ImageNet initialized models, it takes about 50 epochs to achieve similar performance.

These findings in this section underscore the importance of employing a comprehensive agricultural dataset like CWD30 for pretraining deep learning models. By utilizing the rich and diverse data offered by CWD30, researchers can develop efficient and generalizable deep learning models that are more suitable for a wide range of applications, including precision agriculture.

